\documentclass[10pt,letterpaper]{article}
\usepackage{preprint}

\usepackage{amsmath,amsfonts,bm}

\def\eqref#1{equation~\ref{#1}}

\def\1{\bm{1}}

\def\va{{\bm{a}}}
\def\vb{{\bm{b}}}
\def\vc{{\bm{c}}}
\def\vd{{\bm{d}}}
\def\ve{{\bm{e}}}

\def\vh{{\bm{h}}}

\def\vk{{\bm{k}}}

\def\vq{{\bm{q}}}

\def\vv{{\bm{v}}}

\def\vx{{\bm{x}}}

\def\mA{{\bm{A}}}

\def\mH{{\bm{H}}}
\def\mI{{\bm{I}}}

\def\mK{{\bm{K}}}

\def\mP{{\bm{P}}}

\def\mW{{\bm{W}}}

\DeclareMathAlphabet{\mathsfit}{\encodingdefault}{\sfdefault}{m}{sl}
\SetMathAlphabet{\mathsfit}{bold}{\encodingdefault}{\sfdefault}{bx}{n}

\DeclareMathOperator*{\argmax}{arg\,max}
\DeclareMathOperator*{\argmin}{arg\,min}

\usepackage{graphicx}
\usepackage{booktabs}
\usepackage{multirow}
\usepackage{float}
\usepackage{wrapfig}
\usepackage{xcolor}
\usepackage{hyperref}
\usepackage{url}
\usepackage{amsmath,amssymb,amsthm}
\usepackage{enumitem}
\usepackage{tikz}
\usepackage[subpreambles=true]{standalone}
\usetikzlibrary{arrows.meta,positioning,fit,calc}
\usepackage[nameinlink,capitalize]{cleveref}

\newenvironment{resulttable}[3][H]
  {\begin{table}[#1]%
   \setlength{\abovecaptionskip}{1pt}\setlength{\belowcaptionskip}{2pt}%
   \centering\small\caption{#3}\label{#2}}
  {\end{table}}
\AtBeginDocument{%
  \setlength{\abovedisplayskip}{7pt plus 2pt minus 2pt}%
  \setlength{\belowdisplayskip}{7pt plus 2pt minus 2pt}%
  \setlength{\abovedisplayshortskip}{4pt plus 1pt minus 1pt}%
  \setlength{\belowdisplayshortskip}{5pt plus 1pt minus 1pt}%
}
\newcommand{\shrinktofit}[1]{%
  \setbox0=\hbox{#1}%
  \ifdim\wd0>\linewidth \resizebox{\linewidth}{!}{\usebox0}\else \usebox0 \fi}
\newtheorem{theorem}{Theorem}
\newtheorem{proposition}[theorem]{Proposition}
\newtheorem{lemma}[theorem]{Lemma}

\newtheorem{remark}{Remark}
\theoremstyle{definition}
\newtheorem{definition}{Definition}
\theoremstyle{plain}

\title{CrossGMN: Graph Metanetworks for Cross-Architecture Weight-Space Transformations}

\author{%
  \textbf{Adir Dayan}\textsuperscript{1}%
  \quad\textbf{Yam Eitan}\textsuperscript{1}%
  \quad\textbf{Haggai Maron}\textsuperscript{1,2}\\[4pt]
  {\small\textsuperscript{1}Technion -- Israel Institute of Technology
  \quad\textsuperscript{2}NVIDIA Research}%
}
\date{}
\hypersetup{
  pdftitle={CrossGMN: Graph Metanetworks for Cross-Architecture Weight-Space Transformations},
  pdfauthor={Adir Dayan, Yam Eitan, Haggai Maron},
  pdfsubject={},
  pdfkeywords={}
}
\ifdefined\pdfinfoomitdate\pdfinfoomitdate=1\fi

\begin{document}
\addtocontents{toc}{\protect\setcounter{tocdepth}{-1}}

\maketitle

\begin{abstract}
Weight-space networks operate directly on parameters of other neural networks, enabling tasks such as predicting model properties, editing trained models, and generating weights. Equivariance has emerged as a key design principle for these networks, owing to weight-space symmetries such as neuron permutations. However, existing equivariant weight-space architectures have primarily been studied for transformations that preserve the network architecture. In contrast, many practical transformations, including model compression and upscaling, map a trained source network into a target network with a different architecture. In this setting, the source and target permutation symmetries act on different parameter spaces, making equivariance less straightforward to formulate.
Our key idea for addressing this mismatch is to reformulate cross-architecture operators with two inputs: a trained source network and an initialization of the target network.
This lets us define equivariant cross-architecture operators that refine the initialization of the target network using information from the source network, while being invariant to source-network permutations and equivariant to target-network permutations. Based on this formulation, we introduce CrossGMN, a graph metanetwork that jointly processes both networks through symmetry-preserving cross-network message passing. We prove CrossGMN is universal for continuous cross-architecture operators on compact sets under a general-position assumption.
We further evaluate CrossGMN for model compression, where it predicts the parameters of a smaller network to accelerate subsequent knowledge distillation. Across 2-D and 3-D INRs and image classification with MLPs, CNNs, and Vision Transformers, CrossGMN speeds up distillation by up to 8.89×, transfers across datasets without retraining (3.78×), and a single model can accelerate compression from heterogeneous source architectures into a common target architecture.
\end{abstract}

\section{Introduction}

Weight-space learning studies neural architectures that operate directly on the parameters of other neural networks, enabling tasks such as predicting model properties, editing trained models, and generating weights \citep{unterthiner2020predicting,eilertsen2020classifying,schurholt2021selfsupervised,schurholt2022hyperrepresentations,schurholt2024saneicml}. A central property of weight spaces is their permutation symmetries, resulting in many different model parameterizations that represent the same function. This has motivated the development of permutation-equivariant weight-space architectures \citep{navon2023equivariant,zhou2023permutation,zhou2023nft,zhou2024universal,lim2024graph,kofinas2024graph,tran2024monomial,tran2025transformer,vo2025polynomial}, which represent the current state of the art for learning over weights in many tasks.

Many practical transformations of neural network weights do not preserve the
network architecture. Model compression maps a trained network into a smaller
one for deployment under memory, latency, or compute constraints, through
structured pruning~\citep{li2017pruning} or knowledge
distillation~\citep{hinton2015distilling,gou2021knowledge}. Cross-architecture
knowledge transfer moves the behavior of a trained network into a structurally
different one \citep{ahn2019variational,liu2024crossarchitecture,czyzewski2021weights}. Model
upscaling initializes a wider or deeper network from a trained smaller one,
allowing training to continue from the transferred parameters rather than from
scratch
\citep{chen2016net2net,wei2016network,chen2021bert2bert,wang2023ligo,pham2024mixturegrowth}.
All of these map the weights of one architecture into the parameter space of
another, and most of them are realized today by a separate optimization process run from
scratch for every source network. 
Learning such a transformation instead would replace this per-network optimization with a
single feedforward pass: one weight-space model is trained once, and every new source network
is then transformed in a fraction of a second.
Existing equivariant weight-space architectures, however, have 
exclusively addressed mappings whose input and output lie in the \emph{same} weight space. The cross-architecture setting is fundamentally different: the
source and target networks carry distinct symmetry groups acting on distinct parameter spaces, so it is not even clear what equivariance should mean.
Designing equivariant cross-architecture weight-space models is thus a central yet overlooked challenge in weight-space learning.

Our key observation is that cross-architecture transformations typically come with a natural
second input. Whether the target is smaller or larger than the source, the
procedure always begins from an \emph{initialized} target network, which is then
optimized against the trained source network, as in knowledge distillation and
its variants~\citep{hinton2015distilling,gou2021knowledge,nasser2024reverse}.
The transformation is therefore a function of two arguments: the trained source
network and an initialization of the target. Importantly, taking both as input resolves
the difficulty above. The initialization specifies the target
architecture and carries the target symmetry group, so the operator has a
well-defined action on each of its arguments: it should be invariant to symmetries of the source network and equivariant to symmetries of the
target. This leads us to study equivariant \emph{cross-architecture operators}
$
F:\mathcal{V}_{\mathrm{src}}\times\mathcal{V}_{\mathrm{tgt}}
\rightarrow\mathcal{V}_{\mathrm{tgt}}.
$

With the right class of operators in place, the remaining question is how to
parameterize it. We introduce \emph{CrossGMN}, a graph metanetwork for learning
such operators. CrossGMN represents each of the two networks as a parameter
graph, joins them with cross edges, and performs message passing over the
resulting joint graph. Information can thus flow between the source and the
target while the two permutation symmetries are respected independently. We
prove that CrossGMN is a universal approximator of continuous equivariant
cross-architecture operators on compact sets under a general-position
assumption, and extend this result to finite collections of source architectures
mapping into a common target weight space.

We evaluate CrossGMN for model compression through knowledge distillation. Given a trained teacher (source network) and an initialization for a smaller student (target network), CrossGMN predicts refined student parameters intended to reduce the number of distillation steps needed to match the teacher.
We evaluate width, depth, and combined compressions across 2-D and 3-D INRs \citep{sitzmann2020siren} and MNIST \citep{lecun1998gradient}/CIFAR-10 \citep{krizhevsky2009learning}
classifiers, on architectures including MLPs, CNNs, and Vision Transformers \citep{dosovitskiy2021image}.
Across these settings, CrossGMN accelerates subsequent distillation for initializing compressed students, achieving speedups of up to $8.89\times$ on MNIST INRs and $6.85\times$ on ModelNet \citep{wu2015modelnet} INRs. CrossGMN also generalizes out of distribution: a model trained on MNIST INRs transfers to FashionMNIST \citep{xiao2017fashion}, achieving up to $3.78\times$ speedup without retraining. Moreover, a single CrossGMN can compress heterogeneous CIFAR-10 MLP teacher architectures into a common smaller student architecture and generalize to unseen teacher architectures.

In summary, our contributions are: (1) formulating cross-architecture weight-space learning as learning equivariant operators on a source network and a target initialization, which makes the
problem's symmetry structure well-defined; (2) introducing CrossGMN, a graph metanetwork for learning equivariant transformations between different neural-network weight spaces; (3) establishing its universality for continuous cross-architecture operators, including settings with multiple source architectures; and (4) demonstrating its effectiveness for model compression, where it accelerates knowledge distillation across diverse architectures and out-of-distribution settings.

\section{Preliminaries}
\label{sec:prelim}

We introduce the notation and background required to define cross-architecture
weight-space operators.
For the remainder of the manuscript, we focus our theoretical analysis on fully connected MLPs, while evaluating our method empirically on a broader range of architectures, including CNNs and Transformers. The corresponding formal treatment for these architectures is provided in Appendix~\ref{app:cnn-graphs}--\ref{app:transformer-graphs}.

\paragraph{Weight spaces.}
An MLP architecture $A=(\vd,\sigma)$ with $L$ layers is
specified by layer widths
$\vd=(d_0,\ldots,d_L)$ and an activation function $\sigma$.
Its parameters are
\[
    \vv
    =
    (\mW_1,\vb_1,\ldots,\mW_L,\vb_L),
    \qquad
    \mW_\ell\in\mathbb R^{d_\ell\times d_{\ell-1}},
    \quad
    \vb_\ell\in\mathbb R^{d_\ell}.
\]
The corresponding realized function
$f_{\vv}:\mathbb R^{d_0}\rightarrow\mathbb R^{d_L}$ is
\[
    f_{\vv}(\vx)
    =
    \mW_L
    \sigma\!\left(
        \cdots
        \sigma(\mW_1\vx+\vb_1)
        \cdots
    \right)
    +\vb_L .
\]

\begin{definition}[Weight space]
\label{def:weight-space}
For an architecture $A=(\vd,\sigma)$, its weight space is
$
    \mathcal V_A
    :=
    \bigoplus_{\ell=1}^{L}
    \left(
        \mathbb R^{d_\ell\times d_{\ell-1}}
        \oplus
        \mathbb R^{d_\ell}
    \right),
$
where, $\oplus$ denotes a direct sum.
\end{definition}


Throughout the paper, we consider a source architecture
$A_{\mathrm{src}}$ and a target architecture
$A_{\mathrm{tgt}}$, with corresponding weight spaces
$\mathcal V_{\mathrm{src}}$ and $\mathcal V_{\mathrm{tgt}}$.
Their hidden widths and depths may differ, but we assume the same input and output dimensions.

\paragraph{Permutation symmetries.}
Permuting hidden neurons changes the parameterization of an MLP while
preserving its realized function. We formalize these transformations through
the following group action.

\begin{definition}[Permutation action on weight space]
\label{def:permutation-action}
For an architecture $A=(\vd,\sigma)$, let
$G_A:=S_{d_1}\times\cdots\times S_{d_{L-1}}$ be its hidden-neuron permutation group.
For $g=(\tau_1,\ldots,\tau_{L-1})\in G_A$, set $\tau_0,\tau_L=e$ and let $\mP_{\tau_\ell}$ denote the corresponding permutation matrix. The action $\rho_A(g):\mathcal V_A\rightarrow\mathcal V_A$ is defined by
$\mW'_\ell=\mP_{\tau_\ell}^{\top}\mW_\ell\mP_{\tau_{\ell-1}}$ and
$\vb'_\ell=\mP_{\tau_\ell}^{\top}\vb_\ell$ for $1\leq \ell\leq L$.
By construction, $f_{\rho_A(g)\vv}=f_{\vv}$ for every $g\in G_A$.
\end{definition}


\begin{definition}[Permutation-invariant and equivariant weight-space networks]
\label{def:invariance-equivariance}
Let $A$ be an architecture, and let $G_A$ act on its weight space
$\mathcal V_A$ through $\rho_A$ as in
Definition~\ref{def:permutation-action}. A weight-space network is a
parameterized map
$
    \Psi_\theta:\mathcal V_A\to Y,
$
where $G_A$ acts on $Y$ through a representation $\rho_Y$.
The network is $G_A$-\emph{equivariant} if
$
    \Psi_\theta(\rho_A(g)\vv)
    = \rho_Y(g)\Psi_\theta(\vv)
$
for all $g\in G_A$ and $\vv\in\mathcal V_A$.
It is $G_A$-\emph{invariant} if
$
    \Psi_\theta(\rho_A(g)\vv)=\Psi_\theta(\vv)
$
for all $g\in G_A$ and $\vv\in\mathcal V_A$.
\end{definition}
In the next section, we consider the
cross-architecture setting in which the source and target networks belong to
different weight spaces with independent permutation symmetries.

\section{Cross-Architecture Weight-Space Transformations}
\label{sec:crossarch}

Let $A_{\mathrm{src}}$ and $A_{\mathrm{tgt}}$ be two architectures with
weight spaces $\mathcal V_{\mathrm{src}}$ and $\mathcal V_{\mathrm{tgt}}$, and
permutation groups $G_{\mathrm{src}}$ and $G_{\mathrm{tgt}}$.

\paragraph{Limitations of predicting the target directly from the source.}
One could directly map source parameters to target parameters,
$
T:\mathcal V_{\mathrm{src}}\longrightarrow\mathcal V_{\mathrm{tgt}}.
$
However, this formulation has two difficulties. First, its equivariance
structure is unclear: because the source and target have different permutation
groups acting on different weight spaces, $T$ must be invariant to source
permutations while selecting a representative among equivalent target
parameterizations.
Second, natural source-to-target maps may be discontinuous and thus cannot be uniformly approximated arbitrarily well by continuous neural networks. Appendix~\ref{sec:counterexample} gives a simple setting of model compression where mapping each larger network to an optimal smaller one is discontinuous.

Instead, we formulate the problem as refinement from an initialization of the
target network; this gives rise to more natural continuous cross-architecture operators.
This leads to the following definition.

\begin{definition}[Cross-architecture weight-space operator]
\label{def:cross-architecture-operator}
A cross-architecture weight-space operator is a map
\begin{equation}
\label{eq:cross-operator}
    F:\mathcal V_{\mathrm{src}}\times\mathcal V_{\mathrm{tgt}}
    \longrightarrow \mathcal V_{\mathrm{tgt}},
    \qquad
    (\vv_{\mathrm{src}},\vv_{\mathrm{init}})
    \longmapsto F(\vv_{\mathrm{src}},\vv_{\mathrm{init}}).
\end{equation}
The second input $\vv_{\mathrm{init}}$ both
specifies a concrete parameterization in the target weight space and provides
the parameters that the operator can refine.
We say
that $F$ is equivariant with respect to the source and target permutations if
\begin{equation}
\label{eq:cross-equivariance}
    F(g_{\mathrm{src}}\!\cdot\vv_{\mathrm{src}},
      g_{\mathrm{tgt}}\!\cdot\vv_{\mathrm{init}})
    =
    g_{\mathrm{tgt}}\!\cdot F(\vv_{\mathrm{src}},\vv_{\mathrm{init}})
    \quad
    \forall (g_{\mathrm{src}},g_{\mathrm{tgt}})\in G_{\mathrm{src}} \times G_{\mathrm{tgt}}
\end{equation}
\end{definition}

A natural and practically important instance of the operator class above is
knowledge distillation~\citep{hinton2015distilling}. In its simplest setting, a
target network, called the \emph{student}, is trained to match the outputs of a
source network, called the \emph{teacher}. In our setting, we denote their
architectures by $A_T=A_{\mathrm{src}}$ and $A_S=A_{\mathrm{tgt}}$, their
weight spaces by $\mathcal V_T$ and $\mathcal V_S$, the trained teacher by
$\vv_T$, and the student by $\vv_S$. Distillation then minimizes
\begin{equation}
\label{eq:functional-loss}
    L(\vv_T,\vv_S)
    :=\mathbb E_{\vx\sim\mathcal D}
    \left[\ell\!\left(f_{\vv_T}(\vx),f_{\vv_S}(\vx)\right)\right],
\end{equation}
where $\ell$ measures the discrepancy between teacher and student outputs.

Let $\Phi_k(\vv_T,\vv_0)$ denote the student parameters obtained after $k$
distillation steps initialized at $\vv_0$. For SGD~\citep{bottou2012sgd} and
Adam~\citep{kingma2015adam} with fixed hyperparameters and minibatches,
$\Phi_k$ is a continuous equivariant cross-architecture operator in the sense of
Definition~\ref{def:cross-architecture-operator} for every fixed $k$, provided
the student loss gradient is jointly continuous. Appendix~\ref{sec:finite-step}
gives the formal statement and proof.

\section{CrossGMN}
\label{sec:crossgmn-section}

To learn the cross-architecture operators defined in the previous section
(Definition~\ref{def:cross-architecture-operator}), we introduce CrossGMN.
Section~\ref{sec:crossgmn} presents its architecture, and
Section~\ref{sec:expressivity} establishes its universality for
continuous cross-architecture operators.

\subsection{Architecture}
\label{sec:crossgmn}

For parameters $\vv$, let $\mathcal P(\vv)=(V,E)$ denote the parameter graph constructed as in GMN~\citep{lim2024graph}. For an MLP, neurons form the nodes, weights are represented as directed edges between consecutive layers, and biases as node features. The full graph construction and feature specification are given in Appendix~\ref{app:mlp-graphs}.

Given source and target MLP architectures, let $\vv_{\mathrm{src}}$ denote the source-network parameters and $\vv_{\mathrm{init}}$ an initialization of the target network. Let
$\mathcal P(\vv_{\mathrm{src}})=(V^{\mathrm{src}},E^{\mathrm{src}})$ and
$\mathcal P(\vv_{\mathrm{init}})=(V^{\mathrm{tgt}},E^{\mathrm{tgt}})$ denote their parameter graphs. We define the \emph{joint graph} as
\begin{equation}
\label{eq:joint-graph}
\mathcal J(\vv_{\mathrm{src}},\vv_{\mathrm{init}})
:=
\bigl(
V^{\mathrm{src}}\sqcup V^{\mathrm{tgt}},\;
E^{\mathrm{src}}\sqcup E^{\mathrm{tgt}}\sqcup E_\times
\bigr),
\end{equation}
where $E_\times$ denotes the set of cross edges. For the joint graph to respect
both symmetries, $E_\times$ must be invariant under the joint action of
$G_{\mathrm{src}}\times G_{\mathrm{tgt}}$ on neurons: whenever
$(u,v)\in E_\times$, also $(g\cdot u,\,h\cdot v)\in E_\times$ for all
$g\in G_{\mathrm{src}}$ and $h\in G_{\mathrm{tgt}}$. We construct $E_\times$ in
two steps. We first determine how the neurons of a source layer are joined to
those of a target layer, which is dictated by this invariance requirement. We
then choose which pairs of layers to join, which is a free design choice.

\textbf{Joining two layers.}
Consider a source hidden layer and a target hidden layer. Since the neurons of
each can be permuted independently, the only nonempty set of edges between them
that is invariant to both permutations is the complete bipartite one, connecting
every source neuron to every target neuron. We call this a \emph{layerwise join},
following the graph join of two disjoint graphs~\citep{cardoso2013join}. Input
and output neurons, in contrast, are indexed by the coordinates of the realized
function's input and output, and are not permuted. Since the source and target
share input and output dimensions, we join them with the sparser
\emph{positionwise join}: the $i$-th source input neuron is connected only to the
$i$-th target input neuron, and likewise for output neurons.

\textbf{Choosing which layers to join.}
Because each layerwise join is invariant on its own, any choice of which source
and target hidden layers to join yields a symmetry-preserving $E_\times$. We call
this choice a \emph{cross-edge policy}. The most permissive policy,
\emph{all-to-all}, joins every source hidden layer to every target hidden layer,
giving $nm$ layerwise joins for networks with $n$ and $m$ hidden layers. To avoid this quadratic growth with depth, we use the following \emph{strided} policy.

\begin{definition}[Strided cross-edge policy]
\label{def:strided-cross-edge-policy}
For two networks with $n\ge m$ hidden layers, the \emph{strided} policy partitions the $n$ hidden layers of the deeper network into $m$
consecutive blocks of near-equal size, and joins the $k$-th hidden layer of the
shallower network to every layer in the $k$-th block. Each hidden layer of the
deeper network is thus joined to exactly one hidden layer of the shallower
network, for a total of $n$ layerwise joins. For example, with $n=10$ and $m=5$,
the $k$-th hidden layer of the shallower network is joined to layers $2k-1$ and
$2k$ of the deeper network, giving $10$ joins instead of the $50$ required by
all-to-all. The formal construction is given in Appendix~\ref{app:cross-policies}.
\end{definition}

Importantly, the strided policy retains the benefits of all-to-all connectivity while using fewer cross edges. In Section~\ref{sec:expressivity}, we establish universality for the strided construction, and our experiments later show that it performs comparably to all-to-all connectivity in practice (Table~\ref{tab:ablations}). We now have all the components needed to define CrossGMN.

\begin{figure}[t]
\centering
\includestandalone[width=0.82\linewidth]{figures/crossgmn}
\caption{\textbf{CrossGMN.} The source parameters $v_{\mathrm{src}}$ and an initialization $v_{\mathrm{init}}$ of the target network are represented as parameter graphs and combined into the joint graph $\mathcal J(v_{\mathrm{src}},v_{\mathrm{init}})$. A GNN operates on the joint graph, and its outputs corresponding to the target network form the parameter update $\bm\Delta_\theta$, yielding the CrossGMN output defined in Definition~\ref{def:crossgmn}.}
\label{fig:crossgmn}
\end{figure}

\begin{definition}[CrossGMN]
\label{def:crossgmn}
Given the joint graph $\mathcal J(\vv_{\mathrm{src}},\vv_{\mathrm{init}})$, let
$\mathrm{GMN}_\theta$ denote a graph metanetwork with learnable parameters
$\theta$ operating on the joint graph, and let $\Pi_{\mathrm{tgt}}$ denote the map that extracts the target-side parameters and arranges them into a vector in the target weight space.
We define the resulting parameter
update as
$$
\bm\Delta_\theta(\vv_{\mathrm{src}},\vv_{\mathrm{init}})
=
\Pi_{\mathrm{tgt}}\!\left(
\mathrm{GMN}_\theta\!\left(
\mathcal J(\vv_{\mathrm{src}},\vv_{\mathrm{init}})
\right)
\right).
$$
The \emph{CrossGMN} operator is then
\begin{equation}
\label{eq:crossgmn-update}
\text{CrossGMN}_\theta(\vv_{\mathrm{src}},\vv_{\mathrm{init}})
=
\vv_{\mathrm{init}}
+
s\,\bm\Delta_\theta(\vv_{\mathrm{src}},\vv_{\mathrm{init}}),
\end{equation}
where $s>0$ is a fixed residual-scale hyperparameter. Figure~\ref{fig:crossgmn} illustrates CrossGMN.
\end{definition}

By construction, CrossGMN is invariant to source permutations and equivariant to target permutations, as required by Definition~\ref{def:cross-architecture-operator}. Message passing on the joint graph preserves both symmetry actions, while $\Pi_{\mathrm{tgt}}$ discards the source and retains the target, yielding the desired invariance and equivariance. A formal statement and proof are given in Proposition~\ref{app:prop-equivariance-families} in Appendix~\ref{app:joint-equivariance}.

\subsection{Expressivity}
\label{sec:expressivity}
\label{sec:crossarch-universality}

Symmetry-preserving architectures restrict the hypothesis space and may therefore limit approximation power, making expressivity analysis particularly important in this setting. We establish that CrossGMN is universal for arbitrary continuous cross-architecture operators satisfying the symmetry in~\eqref{eq:cross-equivariance}. Our result applies whenever the cross-edge connectivity contains all edges of the strided policy in Definition~\ref{def:strided-cross-edge-policy}.

\begin{theorem}[Universality of CrossGMN]
\label{thm:crossarch-universality}
Let $A_{\mathrm{src}}$ and $A_{\mathrm{tgt}}$ be fixed MLP architectures, and let
$K_{\mathrm{src}}\subset\mathcal V_{A_{\mathrm{src}}}$ and
$K_{\mathrm{tgt}}\subset\mathcal V_{A_{\mathrm{tgt}}}$ be compact sets of weights in general position.\footnote{
We use the same general-position assumption as
\citet{dayan2026expressive}; for an MLP, this means that within each hidden layer,
no two neurons have the same bias value.}
Let
$
F:K_{\mathrm{src}}\times K_{\mathrm{tgt}}
\rightarrow \mathcal V_{\mathrm{tgt}}
$
be a continuous cross-architecture operator in the sense of Definition~\ref{def:cross-architecture-operator}.
Then, for every $\varepsilon>0$, there exists a $\text{CrossGMN}_\theta$ such that
\[
    \sup_{(\vv_{\mathrm{src}},\vv_{\mathrm{init}})
        \in K_{\mathrm{src}}\times K_{\mathrm{tgt}}}
    \left\|
        \text{CrossGMN}_\theta(\vv_{\mathrm{src}},\vv_{\mathrm{init}})
        -F(\vv_{\mathrm{src}},\vv_{\mathrm{init}})
    \right\|
    <\varepsilon.
\]
\end{theorem}

In particular, this theorem provides an approximation guarantee for
CrossGMN applied to the knowledge distillation operator $\Phi_k$
introduced in Section~\ref{sec:crossarch}.

\textbf{Proof idea.}
The proof reduces the cross-architecture setting to the fixed-architecture universality results of \citet{dayan2026expressive}, who showed that prominent permutation-equivariant weight-space architectures, including DWSNets~\citep{navon2023equivariant} and GMNs~\citep{lim2024graph}, have equivalent expressive power and are universal on compact general-position sets for invariant functionals and equivariant operators. Specifically, fixing $\vv_{\mathrm{init}}$ makes $\vv_{\mathrm{src}}\mapsto F(\vv_{\mathrm{src}},\vv_{\mathrm{init}})$ source-invariant, while fixing $\vv_{\mathrm{src}}$ makes $\vv_{\mathrm{init}}\mapsto F(\vv_{\mathrm{src}},\vv_{\mathrm{init}})$ target-equivariant. We approximate these components using the corresponding invariant and equivariant universality results and combine them through the cross edges. The full proof is given in Appendix~\ref{sec:universality}.

\paragraph{A single CrossGMN for heterogeneous architectures.}
\label{rem:multi-source-universality}
The universality result extends to any finite collection of source architectures
$A_{\mathrm{src}}^{(1)},\ldots,A_{\mathrm{src}}^{(m)}$ sharing a common target
architecture $A_{\mathrm{tgt}}$, provided the source architecture is identifiable
from the joint graph. The formal statement and proof are given in
Appendix~\ref{sec:any-dim}, and we evaluate this setting empirically in
Section~\ref{sec:multi-architecture}.

\begin{resulttable}[t]{tab:inr-main}{\textbf{INR compression.} We report median speedup + IQR (in brackets) relative to the best pruning-based anchor initialization. The last two columns report the percentage of held-out teachers for which CrossGMN accelerates distillation, and the percentage for which CrossGMN fails to reach the target within the budget.}
\shrinktofit{%
\begin{tabular}{llllrr}
\toprule
Dataset & Reduction axis & Architecture change & \multicolumn{1}{l}{Speedup} & $>1\!\times$ (\%) & Failed (\%) \\
\midrule
MNIST & Width & $[2,32,32,1]\!\to\![2,16,16,1]$ & $2.20\!\times_{\scriptscriptstyle[1.2,\,3.9]}$ & 83 & 2 \\
& Width & $[2,32,32,1]\!\to\![2,8,8,1]$ & $5.20\!\times_{\scriptscriptstyle[1.1,\,12.5]}$ & 76 & 18 \\
& Width & $[2,32,32,1]\!\to\![2,4,4,1]$ & $4.10\!\times_{\scriptscriptstyle[0.0,\,14.1]}$ & 62 & 35 \\
& Depth & $[2,32,32,1]\!\to\![2,32,1]$ & $4.07\!\times_{\scriptscriptstyle[2.2,\,6.6]}$ & 90 & 7 \\
& Width+depth & $[2,32,32,1]\!\to\![2,16,1]$ & $8.89\!\times_{\scriptscriptstyle[3.6,\,16.8]}$ & 86 & 13 \\
\midrule
ModelNet40 & Width & $[3,48,48,48,48,1]\!\to\![3,24,24,24,24,1]$ & $1.11\!\times_{\scriptscriptstyle[0.9,\,1.7]}$ & 57 & 18 \\
& Depth & $[3,48,48,48,48,1]\!\to\![3,48,48,1]$ & $6.85\!\times_{\scriptscriptstyle[1.0,\,19.8]}$ & 74 & 12 \\
& Depth & $[3,48,48,48,48,1]\!\to\![3,48,1]$ & $2.88\!\times_{\scriptscriptstyle[2.0,\,4.0]}$ & 96 & 0 \\
& Width+depth & $[3,48,48,48,48,1]\!\to\![3,24,24,1]$ & $2.46\!\times_{\scriptscriptstyle[0.4,\,4.8]}$ & 68 & 23 \\
\bottomrule
\end{tabular}}
\end{resulttable}

\section{Experiments}
\label{sec:experiments}

We evaluate CrossGMN for model compression through knowledge distillation, a practical instance of the cross-architecture operators discussed in Section~\ref{sec:crossarch}. Following that example, we refer to the source network as the teacher and the target network as the student. We measure whether student initializations predicted by CrossGMN accelerate subsequent distillation toward their teachers.
Our experiments cover several model families and compression settings, including transfer across datasets and to unseen source architectures. Full experimental details are provided in Appendix~\ref{app:core}--\ref{app:cnn}.

\subsection{Experimental setup and evaluation}
\label{sec:exp_setup}

\paragraph{Training and evaluation protocol.}
For each compression setting, we train CrossGMN on a population of trained teachers and student initializations using the distillation objective in~\eqref{eq:functional-loss}. Given a teacher $\vv_T$, CrossGMN predicts
$\hat{\vv_S}=\text{CrossGMN}_{\theta}(\vv_T,\vv_0)$,
where $\vv_0$ is an initialization of the target network as defined in Section~\ref{sec:crossarch}, which we refer to as the \emph{anchor} throughout the compression experiments.

We initialize the anchor $\vv_0$ by applying structured pruning methods to the teacher $\vv_T$, as this provides a stronger starting point for knowledge distillation than random initialization with little additional compression overhead~\citep{xu2024initializing}.
For each experiment, we select the structured-pruning method that yields the lowest loss at initialization among the evaluated methods, providing a strong anchor for CrossGMN to improve upon.
We compare CrossGMN with the selected anchor itself and with other structured-pruning baselines, including magnitude-based~\citep{xu2024initializing}, activation-based~\citep{hu2016network}, and OBS/OBC pruning~\citep{hassibi1992obs,frantar2022obc}.
Appendix~\ref{app:student-init} details the anchor and baseline initialization methods.

At evaluation, we measure the distillation speedup of CrossGMN relative to the anchor separately for each held-out teacher.
We first distill the anchor $\vv_0$ until its relevant validation metric plateaus,\footnote{We define a plateau as no improvement greater than $\epsilon=10^{-4}$ for $P=50$ consecutive evaluations, following the standard early-stopping principle~\citep{prechelt1998automatic}.} and denote the attained value by $q_{\vv_T}^\star$. We then distill the CrossGMN prediction $\hat{\vv}_S$ toward the same teacher under a fixed budget. The \emph{per-teacher speedup} $S_{\vv_T}$ is the ratio between the numbers of distillation steps required by $\vv_0$ and $\hat{\vv}_S$ to reach $q_{\vv_T}^\star$. Thus, $S_{\vv_T}>1$ indicates that CrossGMN accelerates distillation relative to the anchor. If $\hat{\vv}_S$ does not reach $q_{\vv_T}^\star$ within the budget, we count the teacher as a failure and set $S_{\vv_T}=0$.
We report the median speedup over all held-out teachers together with two complementary statistics: the acceleration rate, defined as the fraction of teachers with $S_{\vv_T}>1$, and the failure rate, defined as the fraction of teachers with $S_{\vv_T}=0$. We additionally report wall-clock speedup, including the cost of the CrossGMN forward pass.

\begin{figure}[t]
\centering
\includegraphics[width=\linewidth]{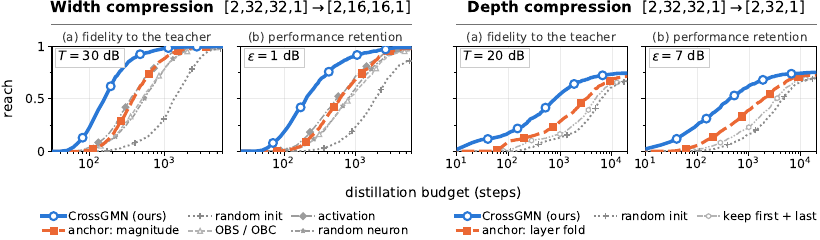}
\caption{\textbf{Reach-budget plots for MNIST INRs.}
Fraction of $1{,}000$ held-out teachers whose students reach a prescribed performance threshold defined relative to the teacher as a function of distillation budget (log scale), comparing CrossGMN with its anchor and other baselines (Appendix~\ref{app:inr}).
Each compression setting includes two panels: \textbf{(a)}~\emph{Fidelity to the teacher}, measured by $\mathrm{PSNR}(\text{student},\text{teacher})\ge T$ dB, and \textbf{(b)}~\emph{Student performance retention}, measured by $\mathrm{PSNR}(\text{student},\text{image})\ge \mathrm{PSNR}(\text{teacher},\text{image})-\varepsilon$. The thresholds $T$ and $\varepsilon$ are chosen separately for each compression setting to account for differences in student architecture capacity. See Section~\ref{sec:exp_setup} for the full description of the reach-budget plots.}
\label{fig:reach-budget-main}
\end{figure}

\paragraph{Reach--budget plots.}
A speedup metric relative to an anchor may be less informative when the anchor itself is weak. We therefore complement speedup metrics with \emph{reach--budget plots}, which evaluate whether the student reaches prescribed performance defined relative to the teacher.
Each plot shows the fraction of teachers whose students reach a fixed target as the distillation budget increases. We report two measures: \emph{fidelity reach}, which tracks how quickly students match their teachers' functions, and \emph{performance-retention reach}, which tracks how quickly they attain a prescribed level of their teachers' task performance. Figure~\ref{fig:reach-budget-main} shows both measures for representative settings; Appendices~\ref{app:protocol} and~\ref{app:reach} provide the full protocol and plots for all settings.

\paragraph{Teacher model families.}
We evaluate compression on trained-model populations~\footnote{To justify amortization, each model population should span diverse functions; classifier diversity is quantified in Appendix~\ref{app:diversity}.}, including
2-D SIREN implicit neural representations (INRs)
\citep{sitzmann2020siren} fitted to MNIST images \citep{lecun1998gradient},
3-D SIREN INRs fitted to ModelNet40 shapes \citep{wu2015modelnet}, MLP and
Vision Transformer (ViT) classifiers \citep{dosovitskiy2021image} on CIFAR-10
\citep{krizhevsky2009learning}, and CNN classifiers on MNIST from the model zoo of \citet{schurholtzoo}.

\subsection{CrossGMN distillation speedup across model families and datasets}
\label{sec:compression-results}

\paragraph{INR datasets.}
For MNIST, we use the dataset of \citet{navon2023equivariant}, containing one SIREN INR per image with teacher architecture $[2,32,32,1]$ ($1{,}185$ parameters). For ModelNet40, we use the SIREN dataset of 3-D shape distance fields from \citet{shamsian2024augmentations}, with teacher architecture $[3,48,48,48,48,1]$ ($7{,}297$ parameters). CrossGMN is trained and evaluated on disjoint teacher subsets for each dataset; Appendix~\ref{app:inr}--\ref{app:modelnet} provides full details. Table~\ref{tab:inr-main} reports distillation speedups across compression settings, and Figure~\ref{fig:reach-budget-main} shows reach--budget curves for representative MNIST width and depth reductions, with wall-clock speedups reported below.

\begin{resulttable}[t]{tab:classifier-main}{\textbf{Classifier compression.} Median distillation speedup + IQR (in brackets) relative to the best pruning-based anchor initialization; columns as defined in Table~\ref{tab:inr-main}.}
\shrinktofit{%
\begin{tabular}{lllrrr}
\toprule
Family & Reduction axis & Architecture change & \multicolumn{1}{l}{Speedup} & $>1\!\times$ (\%) & Failed (\%) \\
\midrule
\multirow{3}{*}{MLP}
& Width & $[3072,64,64,64,64,10]\!\to\![3072,32,32,32,32,10]$
& $1.04\!\times_{\scriptscriptstyle[0.8,\,1.4]}$ & 54 & 5 \\
& Depth & $[3072,64,64,64,64,10]\!\to\![3072,64,64,10]$
& $1.25\!\times_{\scriptscriptstyle[0.7,\,2.3]}$ & 60 & 3 \\
& Width+depth & $[3072,64,64,64,64,10]\!\to\![3072,32,32,10]$
& $1.18\!\times_{\scriptscriptstyle[0.8,\,1.8]}$ & 59 & 3 \\
\midrule
\multirow{4}{*}{ViT}
& Depth & blocks $4\!\to\!2$
& $1.21\!\times_{\scriptscriptstyle[1.1,\,1.3]}$ & 95 & 0 \\
& Attention dimension & $48\!\to\!24$
& $0.74\!\times_{\scriptscriptstyle[0.6,\,0.9]}$ & 7 & 5 \\
& Head count & heads $2\!\to\!1$
& $0.86\!\times_{\scriptscriptstyle[0.8,\,1.0]}$ & 16 & 0 \\
& Head count+depth & heads $2\!\to\!1$, blocks $4\!\to\!2$
& $1.18\!\times_{\scriptscriptstyle[1.1,\,1.3]}$ & 92 & 0 \\
\midrule
CNN
& Width & conv $[8,6,4]\!\to\![4,3,2]$
& $1.17\!\times_{\scriptscriptstyle[0.9,\,1.4]}$ & 68 & 1 \\
\bottomrule
\end{tabular}}
\end{resulttable}

\paragraph{Wall-clock speedup.}
We additionally report wall-clock speedup, including graph construction and the CrossGMN forward pass. For the MNIST INR settings in Table~\ref{tab:inr-main}, CrossGMN inference takes $18$--$20$\,ms per teacher on an NVIDIA A100, compared with $1.8$--$2.4$\,ms per distillation step. This overhead is small relative to the overall distillation budget: width compression from $[2,32,32,1]$ to $[2,8,8,1]$, with a budget of $2{,}480$ steps, achieves a $4.99\times$ wall-clock speedup versus $5.20\times$ in distillation steps, while joint width--depth compression to $[2,16,1]$, with a budget of $2{,}750$ steps, achieves $8.36\times$ versus $8.89\times$. Full timing details and results are provided in Appendices~\ref{app:protocol} and~\ref{app:full}.

\paragraph{Classifier datasets.}
We conduct separate compression experiments on MLP, ViT, and CNN model zoos. For MLPs and ViTs, we construct separate datasets of models trained on CIFAR-10, with $5{,}000$ training and $1{,}000$ test models per dataset. The MLPs have architecture $[3072,64,64,64,64,10]$ ($209{,}802$ parameters), while the ViTs have four Transformer blocks, two attention heads, hidden dimension $48$, and FFN width $192$ ($115{,}930$ parameters). For CNNs, we use the existing model zoo of \citet{schurholtzoo}, containing $2{,}560$ MNIST classifiers with three convolutional layers of widths $[8,6,4]$ ($2{,}464$ parameters). Table~\ref{tab:classifier-main} reports the compression settings and results; Appendix~\ref{app:mlp}--\ref{app:cnn} provides dataset construction and training details.

\paragraph{Heterogeneous-architecture compression.}
When a model collection spans multiple architectures, training a separate compression operator for each reduces the benefits of amortization. To evaluate this setting, we train a single CrossGMN on four CIFAR-10 MLP teacher architectures, all mapped to a common student architecture. Table~\ref{tab:multi-architecture} reports results for held-out teachers from each architecture and for the pooled population.

\label{sec:multi-architecture}
\begin{wraptable}{r}{0.485\linewidth}
\centering\small
\setlength{\abovecaptionskip}{0pt}\setlength{\belowcaptionskip}{0pt}
\caption{\textbf{Heterogeneous-architecture compression.} A single CrossGMN maps all four source architectures to $[3072,8,10]$.}
\label{tab:multi-architecture}
\setlength{\tabcolsep}{2pt}
\renewcommand{\arraystretch}{0.95}
\shrinktofit{%
\begin{tabular}{lrrr}
\toprule
Teacher & \multicolumn{1}{l}{Speedup} & $>1\!\times$ (\%) & Failed (\%) \\
\midrule
$[3072,16,10]$ & $1.28\!\times_{\scriptscriptstyle[0.7,\,1.9]}$ & 60 & 12 \\
$[3072,16,16,10]$ & $1.07\!\times_{\scriptscriptstyle[0.7,\,1.6]}$ & 55 & 14 \\
$[3072,32,10]$ & $1.40\!\times_{\scriptscriptstyle[0.8,\,3.4]}$ & 65 & 6 \\
$[3072,32,32,10]$ & $1.54\!\times_{\scriptscriptstyle[0.7,\,2.9]}$ & 66 & 10 \\
\midrule
Pooled & $1.28\!\times_{\scriptscriptstyle[0.7,\,2.3]}$ & 62 & 10 \\
\bottomrule
\end{tabular}}
\end{wraptable}

\paragraph{Discussion.}
As seen in Table~\ref{tab:inr-main}, CrossGMN consistently accelerates distillation across INRs, achieving significant speedups.
For classifiers, Table~\ref{tab:classifier-main} shows more modest gains that depend on the compression setting: CrossGMN improves distillation for MLPs, CNNs and the ViT depth reductions, while reducing the ViT attention dimension or head count does not yield a speedup.
The results in Table~\ref{tab:multi-architecture} show that CrossGMN achieves
positive speedups across heterogeneous architectures.
Taken together, the results provide evidence that model compression can be learned and amortized directly in weight space, and suggest that CrossGMN is a suitable architecture for studying such learned cross-architecture transformations.

\subsection{Out-of-Distribution Generalization}
\label{sec:cross_task_ood}

A desirable property of weight-space models is to become increasingly agnostic to the underlying network architecture, rather than requiring retraining for each specific model family. We therefore evaluate CrossGMN under two out-of-distribution shifts: cross-dataset transfer and unseen source architectures. Table~\ref{tab:ood-task} reports cross-dataset transfer, while Table~\ref{tab:ood-architecture} evaluates unseen architectures. These results support CrossGMN’s ability to transfer across datasets and provide evidence of partial generalization to unseen architectures.

\begin{table}[H]
\setlength{\abovecaptionskip}{1pt}\setlength{\belowcaptionskip}{2pt}
\centering
\begin{minipage}[t]{0.485\linewidth}
\vspace{0pt}\centering\small
\caption{\textbf{Transfer across datasets.} The CrossGMN models trained on the MNIST INR settings of Table~\ref{tab:inr-main} are applied unchanged to FashionMNIST INRs of the same architectures.}
\label{tab:ood-task}
\setlength{\tabcolsep}{2pt}
\renewcommand{\arraystretch}{1.11}
\shrinktofit{%
\begin{tabular}{lrrr}
\toprule
Architecture change & \multicolumn{1}{l}{Speedup} & $>1\!\times$ (\%) & Failed (\%) \\
\midrule
$[2,32,32,1]\!\to\![2,8,8,1]$ & $2.18\!\times_{\scriptscriptstyle[0.0,\,5.8]}$ & 61 & 33 \\
$[2,32,32,1]\!\to\![2,32,1]$ & $3.09\!\times_{\scriptscriptstyle[1.5,\,6.1]}$ & 86 & 6 \\
$[2,32,32,1]\!\to\![2,16,1]$ & {\unboldmath$3.78\!\times_{\scriptscriptstyle[1.1,\,10.3]}$} & 76 & 17 \\
\bottomrule
\end{tabular}}
\end{minipage}\hfill
\begin{minipage}[t]{0.485\linewidth}
\vspace{0pt}\centering\small
\caption{\textbf{Generalization to unseen architectures.} The CrossGMN checkpoint used in Section~\ref{sec:multi-architecture} is applied without retraining to unseen wider and deeper teacher architectures.}
\label{tab:ood-architecture}
\setlength{\tabcolsep}{2pt}
\shrinktofit{%
\begin{tabular}{lrrr}
\toprule
Source architecture & \multicolumn{1}{l}{Speedup} & $>1\!\times$ (\%) & Failed (\%) \\
\midrule
$[3072,64,10]$ & $1.38\!\times_{\scriptscriptstyle[0.5,\,3.2]}$ & 61 & 11 \\
$[3072,16,16,16,16,10]$ & $0.87\!\times_{\scriptscriptstyle[0.4,\,1.3]}$ & 41 & 17 \\
\midrule
Pooled & $1.06\!\times_{\scriptscriptstyle[0.5,\,1.8]}$ & 51 & 14 \\
\bottomrule
\end{tabular}}
\end{minipage}
\end{table}

\vspace{-5pt}

\subsection{Ablations: the role of cross edges}
\label{sec:exp-ablations}

\begin{wraptable}{r}{0.485\linewidth}
\centering\small
\setlength{\abovecaptionskip}{1pt}\setlength{\belowcaptionskip}{2pt}
\caption{\textbf{Teacher--student communication.} CrossGMN ablations on MNIST INR width reduction $[2,32,32,1]\!\to\![2,8,8,1]$, same compression setting as in Table~\ref{tab:inr-main}.}
\label{tab:ablations}
\setlength{\tabcolsep}{2pt}
\shrinktofit{%
\begin{tabular}{lrrr}
\toprule
Variant & \multicolumn{1}{l}{Speedup} & $>1\!\times$ (\%) & Failed (\%) \\
\midrule
Strided (Definition~\ref{def:strided-cross-edge-policy}) & $5.20\!\times_{\scriptscriptstyle[1.1,\,12.5]}$ & 76 & 18 \\
All-to-all (Section~\ref{sec:crossgmn}) & $5.13\!\times_{\scriptscriptstyle[0.9,\,11.6]}$ & 75 & 18 \\
No cross edges & $3.18\!\times_{\scriptscriptstyle[0.4,\,8.6]}$ & 67 & 24 \\
FiLM conditioning & $3.22\!\times_{\scriptscriptstyle[0.4,\,9.1]}$ & 68 & 24 \\
\bottomrule
\end{tabular}}
\end{wraptable}
We ablate the main architectural and training choices of CrossGMN, including teacher--student communication, cross-edge policy, student initialization, supervision objective, teacher-population size, weight-space backbone, and operator capacity. Table~\ref{tab:ablations} focuses on the communication ablation, which highlights two main findings: removing cross edges or replacing them with FiLM conditioning on a pooled teacher summary~\citep{perez2018film} reduces distillation speedup, supporting direct equivariant neuron-level communication; and the strided policy of Definition~\ref{def:strided-cross-edge-policy} performs comparably to all-to-all connectivity while using fewer cross edges. The remaining ablations are reported in Appendix~\ref{app:ablations-main}.

\section{Related Work}
\label{sec:related-work}

Weight-space learning treats neural-network parameters as structured data
\citep{han2026weightspace}. The pioneering works that started exploring this direction are \cite{eilertsen2020classifying,unterthiner2020predicting,schurholt2021selfsupervised}.   Later approaches use symmetry considerations and include equivariant weight-space
networks~\citep{navon2023equivariant,zhou2023permutation,zhou2023nft,tran2025transformer} and
graph-based models~\citep{lim2024graph,kofinas2024graph}, with extensions to
additional symmetries~\citep{kalogeropoulos2024scalegmn,tran2024monomial}.
Importantly, these symmetry-aware architectures operate within a single weight space, whereas we study
transformations between distinct source and target weight spaces. Related
parameter-generation methods either condition on a target architecture
\citep{zhang2019graph,knyazev2021parameter,liu2019metapruning} or learn
distributions over trained-model populations
\citep{schurholt2022hyperrepresentations,schurholt2024saneicml,erkoc2023hyperdiffusion,
wang2024neural,soro2024diffusion}. They generate new weights rather than
transforming a particular trained network into another weight space. An
extended discussion is provided in Appendix~\ref{app:related-work}.


\section{Conclusion}
\label{sec:conclusion}

We explored learning transformations between the weight spaces of different neural architectures and introduced CrossGMN, an equivariant framework for learning such cross-architecture operators. CrossGMN is a graph metanetwork that operates jointly over the source and target weight spaces, using a trained source network to refine an initialization of the target network. We proved universality for continuous cross-architecture operators and evaluated the framework on model compression with knowledge distillation. Across INRs and classifiers, CrossGMN reduces subsequent distillation steps for both width and depth compression, while supporting reuse across source architectures, transfer across datasets, and partial generalization to unseen source architectures. These results show that equivariant models can learn architecture-changing transformations directly in weight space.

\paragraph{Limitations and future work.}
Our current setting requires access to source parameters, assumes compatible input and output spaces, and is evaluated only at moderate scale (up to roughly $200$k parameters). Scaling to larger models may require sparse or blockwise joint graphs and more efficient cross-network communication. Future work could also extend the operator view beyond compression to settings such as model upscaling and cross-family transfer.

\section*{Acknowledgments}
HM is supported by the Israel Science Foundation through a personal grant
(ISF 264/23) and an equipment grant (ISF 532/23), and by the Career Advancement
Chairs in Artificial Intelligence -- Schmidt Futures.

\bibliography{references}
\bibliographystyle{references}

\appendix

\clearpage
\addtocontents{toc}{\protect\setcounter{tocdepth}{3}}
\begingroup
\renewcommand{\contentsname}{Appendix Contents}
\small
\tableofcontents
\endgroup
\clearpage

\section{Extended Related Work}
\label{app:related-work}

\paragraph{Weight-space learning.}
Weight-space learning treats neural-network parameters as structured data for
learning, representation, and manipulation~\citep{han2026weightspace}.
DWSNets~\citep{navon2023equivariant} and Neural Functional
Networks~\citep{zhou2023permutation} construct permutation-equivariant layers
directly in weight space, while Neural Functional
Transformers~\citep{zhou2023nft} use attention. Graph
Metanetworks~\citep{lim2024graph} and Neural Graph
models~\citep{kofinas2024graph} instead represent networks as graphs and apply
message passing. Related work also considers additional
symmetries~\citep{kalogeropoulos2024scalegmn,tran2024monomial}, structured
parameterizations such as LoRA weights~\citep{yang2025weightspace}, and related
neural artifacts such as gradients~\citep{gelberg2025gradmetanet} and
activations~\citep{barshalom2025actvit}. Universal Neural Functionals extend
equivariant weight-space models to general architectures
\citep{zhou2024universal}, while related approaches extend neural functionals
to Transformers and polynomial weight-space features
\citep{tran2025transformer,vo2025polynomial}. These models have also been used to learn operators on trained networks,
including optimization~\citep{kofinas2024graph} and pruning-oriented
editing~\citep{liu2025metapruning}. Existing expressivity results
compare such architectures over a single weight space
\citep{dayan2026expressive}. In contrast, CrossGMN studies learned operators
between distinct source and target weight spaces with independent permutation
actions.

\paragraph{Parameter generation.}
Parameter-generation methods amortize weight generation across architectures,
tasks, or model populations. Graph HyperNetworks
\citep{zhang2019graph,knyazev2021parameter} and
MetaPruning~\citep{liu2019metapruning} take a target architecture as input and
predict its weights. Hyper-representations
\citep{schurholt2022hyperrepresentations,schurholt2024saneicml} and diffusion-based generators
\citep{erkoc2023hyperdiffusion,wang2024neural,soro2024diffusion} instead learn
a distribution over a population of trained models and sample new weights,
optionally conditioned on the task or dataset. These methods generate weights
for an architecture or population rather than transforming a particular
trained source network from one weight space into another.

\paragraph{Source-conditioned architecture changes.}
Net2Net provides analytic function-preserving transformations that widen or
deepen a network~\citep{chen2016net2net}, and model-growth methods extend this
approach to pretrained Transformers~\citep{chen2021bert2bert}.
LiGO~\citep{wang2023ligo} learns a structured linear map from a smaller
pretrained model to an initialization of a larger one.  While these methods
use analytic growth rules or optimize a growth map for a given pretrained
source model, CrossGMN learns a shared operator across a population of
source networks and reuses it on unseen sources without fitting a new
operator for each model.

\paragraph{Compression and distillation.}
Structured pruning constructs a smaller network by removing entire units or
channels and reusing the retained weights~\citep{li2017pruning}.  Knowledge
distillation trains a student to match a teacher's outputs
\citep{hinton2015distilling,gou2021knowledge}.  FitNets additionally uses
intermediate teacher representations as hints to train thinner, deeper
students~\citep{romero2015fitnets}.  Distillation can also start from weights
selected from the teacher~\citep{xu2024initializing}.  CrossGMN fits between
these stages: it jointly processes the teacher and the structured-pruning
initialization (the anchor) to predict a refined student initialization,
from which distillation proceeds toward the same teacher.  The refinement
is learned across a population of teachers to accelerate this subsequent
optimization.

\section{Notation and Architecture-Specific Symmetries}
\label{app:background}

This section records the notation shared by the theoretical and experimental
appendices and makes precise which permutation actions are respected by the
parameter graphs used for MLPs, CNNs, and Transformers.  The main universality
theorem is stated for MLPs; the final paragraph of Appendix~\ref{app:theory}
explains what transfers to the other families.

\subsection{Notation}
\label{app:notation}

The theoretical sections use the terms \emph{source} and
\emph{target}, whereas the compression with knowledge distillation experiments use \emph{teacher} and
\emph{student}.  Table~\ref{tab:notation} fixes the correspondence.  We write
$g\!\cdot\!\vv$ for the action $\rho_A(g)\vv$ of
Definition~\ref{def:permutation-action}.  In the experiments, the
\emph{anchor} is simply the target initialization supplied to the operator.

\begin{table}[h]
\centering
\small
\caption{Correspondence between the theoretical and experimental notation.}
\label{tab:notation}
\shrinktofit{%
\begin{tabular}{lll}
\toprule
Object & Theory & Compression experiments \\
\midrule
Architectures & $A_{\mathrm{src}},A_{\mathrm{tgt}}$ & teacher $A_T$, student $A_S$ \\
Weight spaces & $\mathcal V_{\mathrm{src}},\mathcal V_{\mathrm{tgt}}$ & $\mathcal V_T,\mathcal V_S$ \\
Trained input & $\vv_{\mathrm{src}}$ & teacher $\vv_T$ \\
Target initialization & $\vv_{\mathrm{init}}$ & anchor $\vv_0$ \\
Operator output & $F(\vv_{\mathrm{src}},\vv_{\mathrm{init}})$ & predicted student $\hat{\vv}_S$ \\
Symmetry groups & $G_{\mathrm{src}},G_{\mathrm{tgt}}$ & $G_T,G_S$ \\
Fixed-step optimizer & $\Phi_k(\vv_{\mathrm{src}},\vv_{\mathrm{init}})$ & $\Phi_k(\vv_T,\vv_0)$ \\
\bottomrule
\end{tabular}}
\end{table}

\subsection{Parameter graphs and induced permutation actions}
\label{app:arch-symmetries}

\paragraph{Parameter graph.}
CrossGMN uses the parameter-graph construction of
\citet{lim2024graph}.  We state only the structure needed for the symmetry
arguments and for reproducing the implementation.

\begin{definition}[Parameter graph]
\label{app:def-param-graph}
For parameters $\vv\in\mathcal V_A$, the parameter graph
$\mathcal P_A(\vv)=(V,E)$ is a typed directed multigraph.  Unit nodes
represent neurons, channels, or attention coordinates.  Auxiliary nodes
represent parameter vectors not indexed by two units, such as biases and the
gain and shift parameters of normalization layers.  Every learnable scalar is
stored on either a node or an edge: a matrix entry $[W]_{ij}$ is carried by the
edge $j\to i$; the entries of a convolutional kernel give parallel edges
between its input and output channels; and bias or normalization entries are
stored on their unit node or on an edge from the corresponding auxiliary node.
A fixed \emph{parameter-to-graph encoding} specifies this assignment and the
correspondence used to decode graph outputs into network parameters
(Appendix~\ref{app:core}).  Nodes carry layer and source/target labels;
classifier nodes also carry node type, canonical input/output indices, and
optional architecture-family and activation-type labels.  Edges carry their
scalar value, layer index, and role (source-internal, target-internal,
source-to-target, or target-to-source in the joint graph), with an optional
architecture-family label for classifiers.
\end{definition}

Following \citet[Section~2.2 and Appendix~C.2]{lim2024graph}, a
\emph{neural DAG automorphism} preserves edge incidence, fixes input, output,
and bias nodes, and preserves weight-sharing constraints.  We use the term
\emph{neural graph automorphism} for the corresponding relabelings of our
parameter and joint graphs, which may include reciprocal edges.  We require
these relabelings to fix auxiliary nodes and preserve all structural node
and edge features, as in \citet[Appendix~C.6]{lim2024graph}.  Parameter values
are permuted with their nodes or edges, rather than held numerically fixed.
Message passing with shared functions is equivariant to these relabelings.
The layerwise joins of Section~\ref{sec:crossgmn} are preserved under
independent relabelings of their two endpoint layers, while positionwise
joins connect only canonically indexed boundary nodes.  Hence
the joint graph preserves the source and target actions independently.

For an MLP, the automorphisms of the standard parameter graph are precisely the
hidden-neuron permutations in Definition~\ref{def:permutation-action}
\citep[Proposition~5]{lim2024graph}.  We first specify the shared cross-edge
policy, then give the full definitions for MLPs, CNNs, and Transformers.

\subsubsection{Joint graphs and cross-edge policies}
\label{app:cross-policies}

\paragraph{Joint graph and layerwise joins.}
For source and target parameter graphs
$\mathcal P_{\mathrm{src}}(\vv_{\mathrm{src}})
=(V^{\mathrm{src}},E^{\mathrm{src}})$ and
$\mathcal P_{\mathrm{tgt}}(\vv_{\mathrm{init}})
=(V^{\mathrm{tgt}},E^{\mathrm{tgt}})$, the joint graph is
\begin{equation}
\label{eq:app-architecture-joint}
\mathcal J(\vv_{\mathrm{src}},\vv_{\mathrm{init}})
=\bigl(V^{\mathrm{src}}\sqcup V^{\mathrm{tgt}},
       E^{\mathrm{src}}\sqcup E^{\mathrm{tgt}}\sqcup E_\times\bigr),
\end{equation}
as in Equation~\ref{eq:joint-graph}.  For node sets $U,W$ on opposite sides,
write
\[
\operatorname{Join}(U,W)=(U\times W)\cup(W\times U)
\]
for the bidirectional layerwise join of Section~\ref{sec:crossgmn}, with each
ordered pair representing a directed cross edge.  Cross-edge structural
features are constant within each directed join.  For CNNs and Transformers,
cross edges have scalar value $0$ and their direction role; auxiliary nodes
retain only their internal connections, as in Appendix~\ref{app:core}.

\paragraph{Positionwise boundary joins.}
For canonically indexed node sets $U=\{u_i:i\in[q]\}$ and
$W=\{w_i:i\in[q]\}$, where $[q]=\{1,\ldots,q\}$, define
\[
\operatorname{Pos}(U,W)
=\{(u_i,w_i),(w_i,u_i):i\in[q]\}.
\]
Source and target share input and output dimensions.  Following
Section~\ref{sec:crossgmn}, their boundary cross edges are
\begin{equation}
\label{eq:app-boundary-cross}
E_{\times,\partial}
=\operatorname{Pos}(U_{\mathrm{in}}^{\mathrm{src}},
                   U_{\mathrm{in}}^{\mathrm{tgt}})
 \cup
 \operatorname{Pos}(U_{\mathrm{out}}^{\mathrm{src}},
                   U_{\mathrm{out}}^{\mathrm{tgt}}).
\end{equation}
These nodes are fixed by the permutation actions, so positionwise joins
preserve the required symmetries.  Internal layer and block boundaries
remain hidden-unit sets and use layerwise joins.

\paragraph{Strided cross-edge policy.}
We give the indexed construction of
Definition~\ref{def:strided-cross-edge-policy}.  Let $n,m\ge0$ be the numbers
of source and target hidden layers, indexed from $0$.  For $n\ge m\ge1$, define
\begin{equation}
\label{eq:app-strided-pairs}
\begin{aligned}
a_i&=\operatorname{round}\!\left(\frac{in}{m}\right),\qquad
b_i=\max\!\left\{\operatorname{round}\!\left(\frac{(i+1)n}{m}\right),
                   a_i+1\right\},\\
\mathcal M_{n,m}
&=\bigcup_{i=0}^{m-1}
  \bigl(\{a_i,\ldots,b_i-1\}\times\{i\}\bigr),
\end{aligned}
\end{equation}
where $\operatorname{round}$ denotes nearest-integer rounding with a fixed
tie convention.  Each target layer $i$ is paired with the source layers in
the nonempty block $\{a_i,\ldots,b_i-1\}$; these blocks partition the source
hidden layers.  For $n<m$, set
$\mathcal M_{n,m}=\{(i,j):(j,i)\in\mathcal M_{m,n}\}$, so pairs always list
the source index first.  If either network has no hidden layers, set
$\mathcal M_{n,m}=\varnothing$.
In particular, $\mathcal M_{n,n}=\{(i,i):0\le i<n\}$ gives the shared-index
policy for width-only compression.  For Transformers, the same matching
rule is applied to blocks, with joins defined below by submodule type.

\subsubsection{Multilayer perceptrons}
\label{app:mlp-graphs}

\paragraph{Weight space.}
Let $A=(\vd,\sigma)$ have widths $\vd=(d_0,\ldots,d_L)$ and entrywise
activation $\sigma$.  As in Definition~\ref{def:weight-space}, its parameters
$\vv=(\mW_1,\vb_1,\ldots,\mW_L,\vb_L)$ lie in
\[
\mathcal V_A
=\bigoplus_{\ell=1}^{L}
  \left(\mathbb R^{d_\ell\times d_{\ell-1}}
        \oplus\mathbb R^{d_\ell}\right).
\]

\paragraph{Realized function.}
For $\vx\in\mathbb R^{d_0}$, set
\[
\vh_0=\vx,\qquad
\vh_\ell=\sigma(\mW_\ell\vh_{\ell-1}+\vb_\ell)
\quad(1\le\ell<L),\qquad
f_\vv(\vx)=\mW_L\vh_{L-1}+\vb_L.
\]
Thus hidden layers apply $\sigma$ and the output layer is affine, as in
Section~\ref{sec:prelim}.

\paragraph{Permutation symmetries.}
The group $G_A=\prod_{\ell=1}^{L-1}S_{d_\ell}$ acts as in
Definition~\ref{def:permutation-action}.  For
$g=(\tau_1,\ldots,\tau_{L-1})$, set
$\mP_{\tau_0}=I_{d_0}$ and $\mP_{\tau_L}=I_{d_L}$, and define
\[
\mW'_\ell=\mP_{\tau_\ell}^{\top}\mW_\ell\mP_{\tau_{\ell-1}},
\qquad
\vb'_\ell=\mP_{\tau_\ell}^{\top}\vb_\ell,
\qquad 1\le\ell\le L.
\]
Since $\sigma$ commutes with permutations, the transformed hidden states
satisfy $\vh'_\ell=\mP_{\tau_\ell}^{\top}\vh_\ell$, giving
$f_{\rho_A(g)\vv}=f_\vv$.  The same permutations relabel the hidden nodes
and their incident edges, while fixing input, output, and auxiliary nodes.
Each layerwise join and positionwise boundary edge is preserved under
independent source and target actions.

\paragraph{Parameter graph.}
The node-bias encoding used in Section~\ref{sec:crossgmn} has one node per
neuron and one directed edge per weight:
\[
\begin{aligned}
U_\ell&=\{u_{\ell,i}:i\in[d_\ell]\},\qquad
V=\bigsqcup_{\ell=0}^{L}U_\ell,\\
E&=\bigsqcup_{\ell=1}^{L}
  \{e_{\ell,ij}:u_{\ell-1,j}\to u_{\ell,i}
    \mid i\in[d_\ell],\,j\in[d_{\ell-1}]\}.
\end{aligned}
\]
The parameter graph $\mathcal P_A(\vv)=(V,E)$ carries scalar features
\[
\omega_\vv(e_{\ell,ij})=(\mW_\ell)_{ij},\qquad
\omega_\vv(u_{\ell,i})=
\begin{cases}
0,&\ell=0,\\
(\vb_\ell)_i,&1\le\ell\le L.
\end{cases}
\]
For the classifier encoding, replace node-stored biases by edges
$e^b_{\ell,i}:\eta_\ell\to u_{\ell,i}$ of value $(\vb_\ell)_i$, adding one
fixed auxiliary node $\eta_\ell$ per affine layer.  Both encodings use the
structural features of Definition~\ref{app:def-param-graph}; their feature
dimensions and reciprocal message-passing edges are specified in
Appendix~\ref{app:core}.

\paragraph{Joint graph.}
For source and target MLPs with $n=L_{\mathrm{src}}-1$ and
$m=L_{\mathrm{tgt}}-1$ hidden layers, use the strided pairs
$\mathcal M_{n,m}$ from Equation~\ref{eq:app-strided-pairs} and the
positionwise boundary edges of Equation~\ref{eq:app-boundary-cross}:
\begin{equation}
\label{eq:app-mlp-cross}
E_\times^{\mathrm{MLP}}
=E_{\times,\partial}\cup
 \bigcup_{(j,i)\in\mathcal M_{n,m}}
 \operatorname{Join}(U_{j+1}^{\mathrm{src}},U_{i+1}^{\mathrm{tgt}}).
\end{equation}
Here $U_{\mathrm{in}}^\alpha=U_0^\alpha$ and
$U_{\mathrm{out}}^\alpha=U_{L_\alpha}^\alpha$ for
$\alpha\in\{\mathrm{src},\mathrm{tgt}\}$.
The joint graph is Equation~\ref{eq:app-architecture-joint} with
$E_\times=E_\times^{\mathrm{MLP}}$; auxiliary bias nodes receive no cross
edges.  The all-to-all policy replaces $\mathcal M_{n,m}$ by
$\{0,\ldots,n-1\}\times\{0,\ldots,m-1\}$.

\subsubsection{Convolutional networks}
\label{app:cnn-graphs}

\paragraph{Weight space.}
Let a convolutional stack have channel counts
$\vc=(c_0,\ldots,c_L)$, kernel sizes $\vk=(k_1,\ldots,k_L)$, and pointwise
nonlinearity $\sigma$.  Its weight space is
\[
\mathcal V_A
=
\bigoplus_{\ell=1}^{L}
\left(
\mathbb R^{c_\ell\times c_{\ell-1}\times k_\ell\times k_\ell}
\oplus
\mathbb R^{c_\ell}
\right).
\]
\paragraph{Realized function.}
For $\vv=(K_1,\vb_1,\ldots,K_L,\vb_L)$, define the affine convolution maps
$\mathcal C_\ell(X)=K_\ell*X+\vb_\ell$, where $*$ denotes spatial convolution
with the architecture's fixed stride and padding, and biases are broadcast
over spatial positions.  The realized function is:
\[
f_\vv(\vx)
=
\bigl(\mathcal C_L\circ\sigma\circ\mathcal C_{L-1}
\circ\cdots\circ\sigma\circ\mathcal C_1\bigr)(\vx).
\]
Fixed channelwise pooling maps may be inserted between layers.  Any dense
layers following the convolutional stack use the MLP parameterization above.

\paragraph{Permutation symmetries.}
The hidden-channel group is
$G_A=\prod_{\ell=1}^{L-1}S_{c_\ell}$.  For
$g=(\tau_1,\ldots,\tau_{L-1})$, with $\tau_0$ and $\tau_L$ the identity,
$\rho_A(g)$ applies $P_{\tau_\ell}^{\top}$ to the output-channel axis and
$P_{\tau_{\ell-1}}$ to the input-channel axis of the $\ell$-th kernel, and
sets $\vb'_\ell=P_{\tau_\ell}^{\top}\vb_\ell$.  Channelwise nonlinearities
and pooling commute with this action; flattening induces the corresponding
permutation of the input coordinates of any following dense layer.  Thus
$f_{\rho_A(g)\vv}=f_\vv$.  In the parameter graph, the same action relabels
the hidden channel nodes and their incident kernel and bias edges.

\paragraph{Parameter graph.}
The corresponding parameter graph $\mathcal P_A(\vv)=(V,E)$ has one node
per channel and one auxiliary bias node per convolutional layer:
\[
U_\ell=\{u_{\ell,i}:i\in[c_\ell]\},\qquad
V=\bigsqcup_{\ell=0}^{L}U_\ell
  \sqcup\{\eta_\ell:\ell\in[L]\},
\]
Its edge set is
\[
\begin{aligned}
E=\bigsqcup_{\ell=1}^{L}\Big(
 &\{e_{\ell,ijrs}:u_{\ell-1,j}\to u_{\ell,i}
       \mid i\in[c_\ell],\,j\in[c_{\ell-1}],\,r,s\in[k_\ell]\}\\
 &\sqcup\{e^b_{\ell,i}:\eta_\ell\to u_{\ell,i}
       \mid i\in[c_\ell]\}\Big),
\end{aligned}
\]
with scalar edge values
\[
\omega_\vv(e_{\ell,ijrs})=(K_\ell)_{ijrs},
\qquad
\omega_\vv(e^b_{\ell,i})=(\vb_\ell)_i.
\]
Thus each input/output channel pair has $k_\ell^2$ parallel kernel edges.
Spatial reuse of a kernel entry does not create additional parameter edges
\citep[Section~2.1.2]{lim2024graph}.  Edge indices specify the
parameter-to-edge correspondence; structural features are those of
Definition~\ref{app:def-param-graph}.  Channelwise activations and pooling
reuse the channel nodes.
If a dense readout follows the stack, continue the layer-node notation as
$U_{L+1},\ldots,U_{L_{\mathrm{net}}}$ using the MLP construction, where
$L_{\mathrm{net}}$ is the total number of convolutional and dense layers.
For the stack alone, $L_{\mathrm{net}}=L$.

\paragraph{Joint graph for width compression.}
Our CNN experiments change widths only, so source and target have the same
sequence of $L_{\mathrm{net}}$ layers.  Using the node sets
$U_\ell^{\mathrm{src}}$ and $U_\ell^{\mathrm{tgt}}$ of their parameter
graphs, join corresponding hidden layers:
\begin{equation}
\label{eq:app-cnn-cross}
E_\times^{\mathrm{CNN}}
  =E_{\times,\partial}\cup
    \bigcup_{\ell=1}^{L_{\mathrm{net}}-1}
       \operatorname{Join}(U_\ell^{\mathrm{src}},U_\ell^{\mathrm{tgt}}).
\end{equation}
This is the equal-depth matching $\mathcal M_{n,n}$ of
Equation~\ref{eq:app-strided-pairs}, with $n=L_{\mathrm{net}}-1$.
The boundary edges $E_{\times,\partial}$ follow
Equation~\ref{eq:app-boundary-cross}, with
$U_{\mathrm{in}}^\alpha=U_0^\alpha$ and
$U_{\mathrm{out}}^\alpha=U_{L_{\mathrm{net}}}^\alpha$ for
$\alpha\in\{\mathrm{src},\mathrm{tgt}\}$.
The CNN joint graph is Equation~\ref{eq:app-architecture-joint} with
$E_\times=E_\times^{\mathrm{CNN}}$.

\subsubsection{Transformers}
\label{app:transformer-graphs}

\paragraph{Weight space.}
Let a Transformer have $L$ blocks, residual width $d$, $H$ heads of dimension
$d_h$, attention width $d_a=Hd_h$, and feed-forward width $d_f$.  For blocks
with biases and two affine LayerNorms, its weight space is
\[
\begin{aligned}
\mathcal V_A
={}&\mathcal V_{\mathrm{emb}}\oplus
\bigoplus_{\ell=0}^{L-1}\Big[
  \big(\mathbb R^{d_a\times d}\oplus\mathbb R^{d_a}\big)^{\oplus3}
  \oplus\big(\mathbb R^{d\times d_a}\oplus\mathbb R^d\big)\\
&\hspace{4em}\oplus
  \big(\mathbb R^{d_f\times d}\oplus\mathbb R^{d_f}\big)
  \oplus\big(\mathbb R^{d\times d_f}\oplus\mathbb R^d\big)
  \oplus\mathbb R^{4d}\Big]
  \oplus\mathcal V_{\mathrm{out}}.
\end{aligned}
\]
The block factors contain the query, key, and value projections, the attention
output projection, the two feed-forward maps, and the LayerNorm gains and
shifts, respectively.  The spaces $\mathcal V_{\mathrm{emb}}$ and
$\mathcal V_{\mathrm{out}}$ contain the trainable input and positional
embeddings (including any class token), and the output head and any final
normalization, respectively.

\paragraph{Realized function.}
To specify the realized function, store tokens as columns of
$X\in\mathbb R^{d\times n}$ and write $\mathbf 1_n$ for the all-ones vector.
Suppressing the block index, the projections for head $h$ are
\[
Q_h=W_{q,h}X+\vb_{q,h}\mathbf 1_n^\top,\qquad
K_h=W_{k,h}X+\vb_{k,h}\mathbf 1_n^\top,\qquad
V_h=W_{v,h}X+\vb_{v,h}\mathbf 1_n^\top.
\]
Here $W_{q,h},W_{k,h},W_{v,h}$ are the $d_h$-row blocks of $W_q,W_k,W_v$.
With softmax applied rowwise, define
\[
\begin{aligned}
T_h(X)&=V_h\,\operatorname{softmax}
  \!\left(Q_h^\top K_h/\sqrt{d_h}\right)^\top,\\
\operatorname{MHA}(X)
  &=W_o\,\operatorname{concat}_{h=1}^{H}T_h(X)+\vb_o\mathbf 1_n^\top,\\
\operatorname{FF}(X)
  &=W_2\,\sigma\!\left(W_1X+\vb_1\mathbf 1_n^\top\right)
    +\vb_2\mathbf 1_n^\top,
\end{aligned}
\]
where concatenation is along the feature dimension and $\sigma$ acts
entrywise.  In the pre-normalization convention, block $\ell$ is
\[
Y=X+\operatorname{MHA}_\ell\!\left(\operatorname{LN}_{\ell,1}(X)\right),
\qquad
B_\ell(X)=Y+\operatorname{FF}_\ell\!\left(\operatorname{LN}_{\ell,2}(Y)\right),
\]
where each LayerNorm acts separately on each token, with learned gain and
shift and a fixed positive stabilizer.  The realized function is
\[
f_\vv(\vx)
=
\bigl(R_\vv\circ B_{L-1}\circ\cdots\circ B_0\circ E_\vv\bigr)(\vx),
\]
where $E_\vv$ embeds the input and adds positional representations, and
$R_\vv$ applies the architecture's readout and any final normalization.

\paragraph{Permutation symmetries.}
We use permutations of coordinates within each head, permutations of heads,
and permutations of feed-forward units, while fixing residual-stream
coordinates:
\[
G_A
=
\prod_{\ell=0}^{L-1}
\left((S_{d_h}\wr S_H)\times S_{d_f}\right).
\]
For $(\tau_\ell,\kappa_\ell)$ in the $\ell$-th factor, the attention action
applies $P_{\tau_\ell}^{\top}$ to the rows and biases of $W_q,W_k,W_v$ and
$P_{\tau_\ell}$ to the columns of $W_o$; the feed-forward action applies
$P_{\kappa_\ell}^{\top}$ to the rows and bias of $W_1$ and
$P_{\kappa_\ell}$ to the columns of $W_2$.  Joint query--key permutations
preserve the attention logits, value permutations are absorbed by $W_o$, and
feed-forward permutations are the usual one-hidden-layer MLP symmetry.
Thus this action preserves the realized function.

This subgroup is compatible with the shared-coordinate attention graph of
\citet[Section~2.1.2 and Figure~3]{lim2024graph}.  They describe attention
edges as ``a set that corresponds to the query, key, and value maps'';
the shared intermediate nodes for these three projections are explicit in
their released graph construction.\footnote{The function
\texttt{self\_attention\_to\_graph} uses the same
\texttt{middle\_neuron\_idx} for the query, key, and value projections and
the input of the output projection; see
\href{https://github.com/cptq/graph_metanetworks/blob/295da2d7b57628d3db2f6f7568f8ebac0ee3389e/gmn/graph_construct/utils.py\#L423-L437}{the released implementation}.}
The subgroup does not exhaust the functional symmetries of Transformers.
For example, standard dot-product attention also permits independent
permutations of query--key and value coordinates within each head, as well
as compensating invertible changes of basis in the query and key projections.

The symmetry implemented by a graph metanetwork is determined by the typed
parameter graph, not only by the computational architecture.  We therefore
distinguish the chosen function-preserving group $G_A$ above from
\[
\widehat G_A
:=
\{g\in G_A:\rho_A(g)\text{ is induced by a neural graph automorphism of }
\mathcal P_A\}.
\]
Whether $\widehat G_A=G_A$ depends on the structural labels preserved by the
graph encoding; fixed head-index labels, for example, can restrict head
permutations.
Our implementation and all formal claims about the deployed Transformer
operator use this represented subgroup $\widehat G_A$, without claiming
equivariance to every functional symmetry.

\paragraph{Parameter graph.}
The parameter graph is built by joining the parameter subgraphs of these
modules.  For block $\ell$, let
\[
\begin{aligned}
R_\ell^s&=\{r_{\ell,i}^s:i\in[d]\}\quad(s=0,1,2),&
A_\ell&=\{a_{\ell,p}:p\in[d_a]\},&
F_\ell&=\{f_{\ell,j}:j\in[d_f]\}.
\end{aligned}
\]
The sets $R_\ell^0,R_\ell^1,R_\ell^2$ index the residual coordinates at the
block input, after attention, and at the block output.  The nodes in
$A_\ell$ jointly index query, key, and value coordinates, with
$p=(h-1)d_h+r$ for coordinate $r$ of head $h$; $F_\ell$ indexes the
feed-forward hidden units.  Add bias nodes
$\mathcal B_\ell=\{\eta_{\ell,t}:t\in\{q,k,v,o,1,2\}\}$ and
normalization nodes
$\mathcal N_\ell=\{\nu_{\ell,s}^{\gamma},\nu_{\ell,s}^{\beta}:s=1,2\}$.
The block node set is
\[
V_\ell=
\Big(\bigsqcup_{s=0}^{2}R_\ell^s\Big)
\sqcup A_\ell\sqcup F_\ell\sqcup\mathcal B_\ell\sqcup\mathcal N_\ell.
\]
Writing $u\xrightarrow{\alpha}v$ for an edge whose scalar value is
$\omega_\vv(e)=\alpha$, the matrix edges are precisely
\[
\begin{aligned}
r_{\ell,i}^0&\xrightarrow{(W_{\ell,t})_{pi}}a_{\ell,p}
  &&(t\in\{q,k,v\},\ i\in[d],\ p\in[d_a]),\\
a_{\ell,p}&\xrightarrow{(W_{\ell,o})_{ip}}r_{\ell,i}^1
  &&(p\in[d_a],\ i\in[d]),\\
r_{\ell,i}^1&\xrightarrow{(W_{\ell,1})_{ji}}f_{\ell,j}
  &&(i\in[d],\ j\in[d_f]),\\
f_{\ell,j}&\xrightarrow{(W_{\ell,2})_{ij}}r_{\ell,i}^2
  &&(j\in[d_f],\ i\in[d]).
\end{aligned}
\]
In particular, each pair $(r_{\ell,i}^0,a_{\ell,p})$ has three parallel
edges, one for each of $W_q,W_k,W_v$.  For every affine map $t$, connect
$\eta_{\ell,t}$ to each output-coordinate node of that map with the
corresponding entry of $\vb_{\ell,t}$.  LayerNorm $s$ adds
two edge families for its gain $\gamma_{\ell,s}\in\mathbb R^d$ and shift
$\beta_{\ell,s}\in\mathbb R^d$:
\[
\nu_{\ell,s}^{\gamma}
  \xrightarrow{(\gamma_{\ell,s})_i}r_{\ell,i}^{s-1},
\qquad
\nu_{\ell,s}^{\beta}
  \xrightarrow{(\beta_{\ell,s})_i}r_{\ell,i}^{s-1}
\qquad(i\in[d],\ s=1,2).
\]
The remaining block edges are the parameter-free residual edges
$r_{\ell,i}^0\xrightarrow{1}r_{\ell,i}^1$ and
$r_{\ell,i}^1\xrightarrow{1}r_{\ell,i}^2$.  These matrix, bias,
normalization, and residual edges together define $E_\ell$.
As in \citet[Section~2.1.2 and Appendix~B]{lim2024graph}, activations,
attention softmax, and tokenwise reuse of the projections introduce no
additional parameter nodes or edges.

The full graph $\mathcal P_A(\vv)=(V,E)$ is the union of these block graphs
and the embedding and readout graphs, identifying
$r_{\ell,i}^2=r_{\ell+1,i}^0$ at consecutive block boundaries.  Patch
embeddings use the convolutional construction above, and linear embeddings
and readouts use the MLP edge rule.  Denote their layer-node sets by
$U_k^{\mathrm{emb}}$ and $U_k^{\mathrm{out}}$, with depths
$L_{\mathrm{emb}}$ and $L_{\mathrm{out}}$, and identify
\[
U_{L_{\mathrm{emb}}}^{\mathrm{emb}}=R_0^0,
\qquad
U_0^{\mathrm{out}}=R_{L-1}^2.
\]
The network's input and output node sets are
$U_{\mathrm{in}}=U_0^{\mathrm{emb}}$ and
$U_{\mathrm{out}}=U_{L_{\mathrm{out}}}^{\mathrm{out}}$.
Any trainable positional or class-token
vector is represented by an auxiliary node with one edge per entry to the
corresponding embedding coordinate; final normalization uses the same
gain/shift construction.  Fixed positional representations introduce no
learnable edges.  The node indices above specify parameter locations;
structural labels and any reciprocal edges used for message passing follow
Definition~\ref{app:def-param-graph} and Appendix~\ref{app:core}.

\paragraph{Joint graph.}
Apply the strided matching $\mathcal M_{n,m}$ of
Equation~\ref{eq:app-strided-pairs} to whole Transformer blocks, with
$n=L_{\mathrm{src}}$ and $m=L_{\mathrm{tgt}}$.  Within each matched block
pair, join corresponding unit sets:
\begin{equation}
\label{eq:app-transformer-strided-cross}
\begin{aligned}
E_{\times,\mathrm{blocks}}
=\bigcup_{(j,i)\in\mathcal M_{n,m}}\Big[
 &\bigcup_{s=0}^{2}
   \operatorname{Join}(R_j^{s,\mathrm{src}},R_i^{s,\mathrm{tgt}})\\
 &{}\cup\operatorname{Join}(A_j^{\mathrm{src}},A_i^{\mathrm{tgt}})
   \cup\operatorname{Join}(F_j^{\mathrm{src}},F_i^{\mathrm{tgt}})
\Big].
\end{aligned}
\end{equation}
Thus joins match attention coordinates, feed-forward hidden units, and
residual coordinates at corresponding submodule boundaries.  Attention joins
include all coordinates on both sides, allowing different head counts.

For matching embedding and readout depths on the two sides, join their
corresponding internal node sets $U_k^{t,\alpha}$ defined above, where
$t\in\{\mathrm{emb},\mathrm{out}\}$ and
$\alpha\in\{\mathrm{src},\mathrm{tgt}\}$.  With the positionwise input/output
edges of Equation~\ref{eq:app-boundary-cross}, the full cross-edge set is
\[
E_\times^{\mathrm{Tr}}
=E_{\times,\partial}\cup E_{\times,\mathrm{blocks}}
 \cup\bigcup_{t\in\{\mathrm{emb},\mathrm{out}\}}\
 \bigcup_{k=1}^{L_t-1}
 \operatorname{Join}(U_k^{t,\mathrm{src}},U_k^{t,\mathrm{tgt}}),
\]
which defines the Transformer joint graph through
Equation~\ref{eq:app-architecture-joint}.  The embedding--block and
block--readout interfaces are already included through
$R_0^{0,\alpha}$ and $R_{L_\alpha-1}^{2,\alpha}$.
For a single patch-embedding layer and linear classifier,
$L_{\mathrm{emb}}=L_{\mathrm{out}}=1$, so the last union is empty.
Set unions include each cross edge
once, even when joins overlap after the block-boundary identifications.

\subsubsection{Equivariance of the joint graph}
\label{app:joint-equivariance}

\begin{proposition}[Equivariance induced by the joint graph]
\label{app:prop-equivariance-families}
Let $\mathcal J$ be the joint graph of Equation~\ref{eq:joint-graph},
with the cross-edge construction of Section~\ref{sec:crossgmn}
(and Appendix~\ref{app:mlp-graphs}) for MLPs,
Appendix~\ref{app:cnn-graphs} for CNNs, or
Appendix~\ref{app:transformer-graphs} for Transformers.
Let $\mathrm{GMN}_\theta$ be a graph metanetwork
as defined by \citet[Section~2.3]{lim2024graph}, and let
$H_{\mathrm{src}}$ and $H_{\mathrm{tgt}}$ be parameter-permutation groups
induced by neural graph automorphisms of the respective parameter graphs.
Then the CrossGMN operator
of Definition~\ref{def:crossgmn} satisfies, for every choice of parameters
$\theta$,
\[
\operatorname{CrossGMN}_\theta
(g_{\mathrm{src}}\!\cdot\vv_{\mathrm{src}},
 g_{\mathrm{tgt}}\!\cdot\vv_{\mathrm{init}})
=
 g_{\mathrm{tgt}}\!\cdot
\operatorname{CrossGMN}_\theta
(\vv_{\mathrm{src}},\vv_{\mathrm{init}})
\]
for all $g_{\mathrm{src}}\in H_{\mathrm{src}}$ and
$g_{\mathrm{tgt}}\in H_{\mathrm{tgt}}$ and all source and target parameters.
For MLPs, we may take $H_{\mathrm{src}}=G_{\mathrm{src}}$ and
$H_{\mathrm{tgt}}=G_{\mathrm{tgt}}$, so this is exactly the symmetry required
by Definition~\ref{def:cross-architecture-operator}.
\end{proposition}

\begin{proof}
Fix $(g_{\mathrm{src}},g_{\mathrm{tgt}})\in
H_{\mathrm{src}}\times H_{\mathrm{tgt}}$ and write
$\mathcal J=\mathcal J(\vv_{\mathrm{src}},\vv_{\mathrm{init}})$.
Let $\gamma$ be the disjoint union of inducing automorphisms of the two
parameter graphs.  Each layerwise join is preserved because its endpoint
sets are preserved; any positionwise boundary edges are fixed.  Cross-edge
features are unchanged within each directed join.  Thus $\gamma$ preserves
the joint graph's structure and structural features, and
\[
\mathcal J(g_{\mathrm{src}}\!\cdot\vv_{\mathrm{src}},
           g_{\mathrm{tgt}}\!\cdot\vv_{\mathrm{init}})
=\gamma\!\cdot\mathcal J.
\]
Graph metanetworks commute with such relabelings
\citep[Appendix~C.6]{lim2024graph}, so
$\mathrm{GMN}_\theta(\gamma\!\cdot\mathcal J)
=\gamma\!\cdot\mathrm{GMN}_\theta(\mathcal J)$.
Moreover, the target projection satisfies
$\Pi_{\mathrm{tgt}}(\gamma\!\cdot Z)
=g_{\mathrm{tgt}}\!\cdot\Pi_{\mathrm{tgt}}(Z)$ for every joint output $Z$:
it discards source entries and decodes target entries into their corresponding
parameter coordinates.  Consequently,
\[
\begin{aligned}
\bm\Delta_\theta(g_{\mathrm{src}}\!\cdot\vv_{\mathrm{src}},
                 g_{\mathrm{tgt}}\!\cdot\vv_{\mathrm{init}})
&=\Pi_{\mathrm{tgt}}\!\left(
  \mathrm{GMN}_\theta(\gamma\!\cdot\mathcal J)\right)\\
&=\Pi_{\mathrm{tgt}}\!\left(
  \gamma\!\cdot\mathrm{GMN}_\theta(\mathcal J)\right)\\
&=g_{\mathrm{tgt}}\!\cdot
  \bm\Delta_\theta(\vv_{\mathrm{src}},\vv_{\mathrm{init}}).
\end{aligned}
\]
Since $g_{\mathrm{tgt}}$ acts linearly on target parameters, adding the
transformed initialization and applying the fixed residual scale $s$ in
Equation~\ref{eq:crossgmn-update} gives the claimed identity.
Taking $g_{\mathrm{tgt}}=e$ gives source invariance, and taking
$g_{\mathrm{src}}=e$ gives target equivariance.
\end{proof}

\providecommand{\relu}{\operatorname{ReLU}}

\section{Theoretical Results}
\label{app:theory}

This appendix proves the three claims used in the main text.  First,
mapping each larger network to an optimal smaller one can be discontinuous
(Appendix~\ref{sec:counterexample}).  Second, any fixed number
of standard distillation steps does define a continuous cross-architecture
operator (Appendix~\ref{sec:finite-step}).  Third, CrossGMN is universal for
continuous source-invariant, target-equivariant operators on compact
general-position sets (Appendix~\ref{sec:universality}), including a finite
collection of source architectures (Appendix~\ref{sec:any-dim}).

\subsection{Discontinuity of mapping a larger network to an optimal smaller one}
\label{sec:counterexample}

\begin{figure}[!t]
\centering
\includegraphics[width=\linewidth]{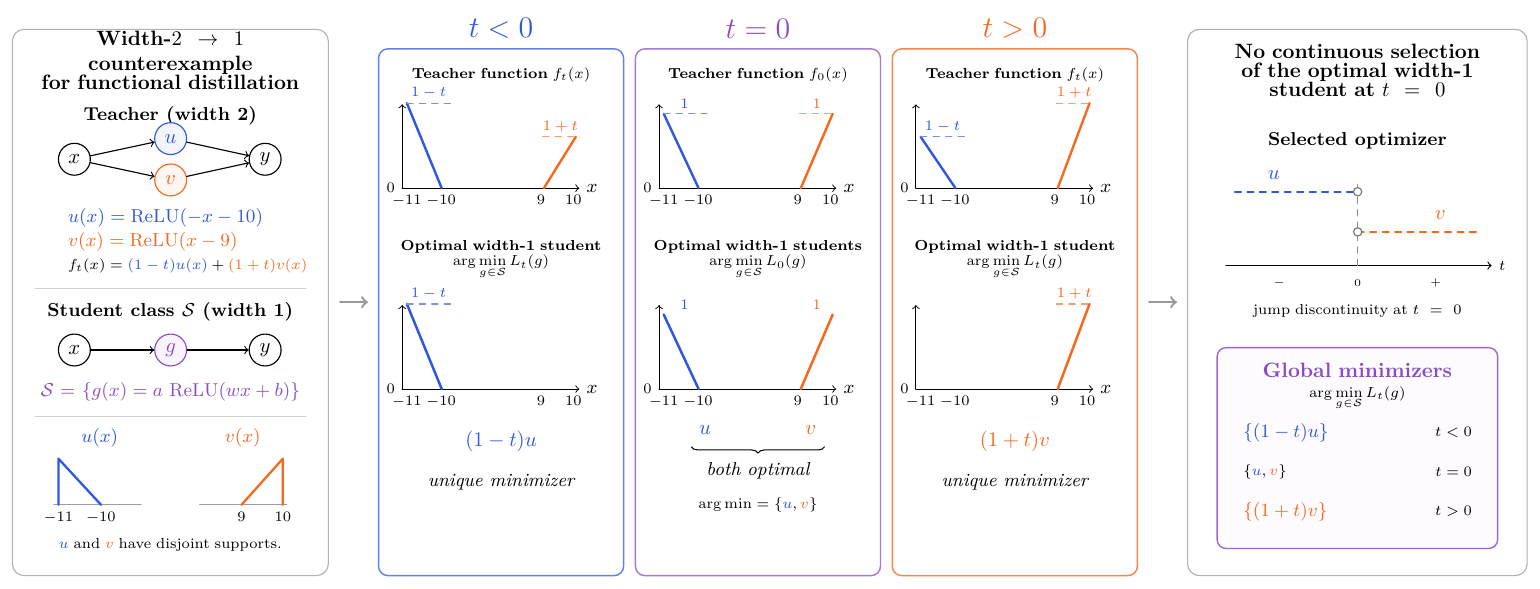}
\caption{\textbf{Mapping a teacher to an optimal student can be discontinuous.}
The width-$2$ teacher parameters $\vv_T(t)$ in
Equation~\ref{eq:source-path} vary continuously and realize
$f_t=(1-t)u+(1+t)v$.  By Lemma~\ref{lem:best-hinge}, the optimal width-$1$
student realizes $(1-t)u$ for $t<0$ and $(1+t)v$ for $t>0$; at $t=0$, both
$u$ and $v$ are optimal.  These one-sided limits differ, so no continuous
parameter map can select an optimal student for every teacher along the
path (Proposition~\ref{prop:counterexample}).}
\label{fig:counterexample}
\end{figure}

In model compression, mapping each larger network to an optimal smaller one
can be discontinuous.  Following Section~\ref{sec:crossarch}, we call the
larger source network the \emph{teacher} and the smaller target network the
\emph{student}, with parameters $\vv_T\in\mathcal V_T$ and
$\vv_S\in\mathcal V_S$.  We consider compressing a ReLU MLP with one hidden layer from width $2$ to width $1$. The optimal student switches between the teacher's two spatially separated ReLU components as their coefficients cross. Figure~\ref{fig:counterexample} illustrates this discontinuity.

\paragraph{Formal construction.}
Take $A_T=((1,2,1),\relu)$ and $A_S=((1,1,1),\relu)$.
On $I=[-11,10]$, define
\begin{equation}
\label{eq:hinges}
 u(x)=\relu(-x-10),
 \qquad
 v(x)=\relu(x-9).
\end{equation}
All norms and inner products below are on $I$.  The supports of $u$ and $v$
are disjoint, and
\begin{equation}
\label{eq:hinge-norms}
 \|u\|_2^2=\|v\|_2^2=\frac13,
 \qquad
 \langle u,v\rangle=0,
 \qquad
 \|u-v\|_\infty=1.
\end{equation}
For $t\in[-\tfrac12,\tfrac12]$, choose teacher parameters $\vv_T(t)$ realizing
\[
 f_t:=f_{\vv_T(t)}=(1-t)u+(1+t)v.
\]
An explicit continuous parameter path is
\begin{equation}
\label{eq:source-path}
 \mW_{T,1}=\begin{bmatrix}-1\\1\end{bmatrix},
 \quad
 \vb_{T,1}=\begin{bmatrix}-10\\-9\end{bmatrix},
 \quad
 \mW_{T,2}(t)=\begin{bmatrix}1-t&1+t\end{bmatrix},
 \quad
 \vb_{T,2}=0.
\end{equation}
Restrict the student to zero output bias:
\[
 \mathcal V_S^0=\{(w,b,a,0):w,b,a\in\mathbb R\}\subset\mathcal V_S,
 \qquad
 f_{\vv_S}(x)=a\relu(wx+b).
\]
Its realized function class is
$\mathcal S=\{f_{\vv_S}|_I:\vv_S\in\mathcal V_S^0\}$, as in
Figure~\ref{fig:counterexample}.  Use squared functional error,
\begin{equation}
\label{eq:counterexample-loss}
 L(\vv_T(t),\vv_S)=L_t(f_{\vv_S}),
 \qquad
 L_t(g):=\int_I |f_t(x)-g(x)|^2\,dx,
\end{equation}
which is the loss of Equation~\ref{eq:functional-loss} for uniform inputs
on $I$, up to a constant factor.  The optimal student parameters are
\begin{equation}
\label{eq:opt-set}
 \operatorname{Opt}(\vv_T(t))
 =\operatorname*{arg\,min}_{\vv_S\in\mathcal V_S^0}
    L(\vv_T(t),\vv_S).
\end{equation}

\begin{lemma}[Best one-hinge approximation]
\label{lem:best-hinge}
The functions realized by the optimal students are
\begin{equation}
\label{eq:opt-characterization}
\{f_{\vv_S}|_I:\vv_S\in\operatorname{Opt}(\vv_T(t))\}
=
\begin{cases}
 \{(1-t)u\}, & t<0,\\[2pt]
 \{u,v\}, & t=0,\\[2pt]
 \{(1+t)v\}, & t>0.
\end{cases}
\end{equation}
\end{lemma}

\begin{proof}
On $I$, every student in $\mathcal V_S^0$ realizes a scalar multiple of
either the constant function, a right-facing hinge
$h_c^+(x)=\relu(x-c)$, or a left-facing hinge
$h_c^-(x)=\relu(c-x)$.  For a nonzero candidate shape $h$, optimizing its
scalar coefficient gives
\begin{equation}
\label{eq:projection-score}
 \min_{\lambda\in\mathbb R}\|f_t-\lambda h\|_2^2
 =\|f_t\|_2^2-\Gamma_t(h),
 \qquad
 \Gamma_t(h)=\frac{\langle f_t,h\rangle^2}{\|h\|_2^2}.
\end{equation}
Thus $\Gamma_t(h)$ is the reduction in squared error obtained by fitting
the best scalar multiple of $h$, relative to the zero function.
Minimizing the student's error is therefore equivalent to maximizing
$\Gamma_t(h)$.

For the teacher's two ReLU components,
\[
 \Gamma_t(u)=\frac{(1-t)^2}{3},
 \qquad
 \Gamma_t(v)=\frac{(1+t)^2}{3}.
\]
For the constant function $h(x)\equiv1$, we have
$\int_I f_t(x)\,dx=(1-t)/2+(1+t)/2=1$ and
$\|h\|_2^2=|I|=21$.  Hence
\[
 \Gamma_t(1)=\frac{1}{21}
 <\frac{1}{12}
 \le\min\{\Gamma_t(u),\Gamma_t(v)\},
\]
where the last inequality follows from $|t|\le1/2$.
Thus the best constant student has strictly larger error than either
optimally scaled component and cannot be optimal.

Consider a nonzero right-facing hinge $h_c^+$.  If $c\ge-10$, it does
not overlap the support of $u$, and Cauchy--Schwarz gives
$\Gamma_t(h_c^+)\le\Gamma_t(v)$, with equality only when
$h_c^+$ is proportional to $v$, i.e. only when $c=9$.  If $c<-10$, put
$\tau=-10-c>0$ and $M_t=\max\{1-t,1+t\}$.  Direct integration gives
\[
 \|h_c^+\|_2^2
 \ge \frac{8000+1200\tau+60\tau^2}{3},
 \qquad
 \langle f_t,h_c^+\rangle
 \le M_t\left(\frac{59}{6}+\tau\right),
\]
and hence
\[
 \Gamma_t(h_c^+)
 \le
 \frac{3M_t^2(59/6+\tau)^2}{8000+1200\tau+60\tau^2}
 <\frac{M_t^2}{3}.
\]
The last inequality is equivalent to
$51\tau^2+1023\tau+28519/4>0$.  Reflecting the interval by
$x\mapsto-1-x$ exchanges $u$ and $v$ and gives the same conclusion for every
left-facing hinge other than $u$.  Therefore the maximizing shape is $u$ for
$t<0$, $v$ for $t>0$, and both for $t=0$.  The minimizing scalar in
\eqref{eq:projection-score} is respectively $1-t$ or $1+t$, proving
\eqref{eq:opt-characterization}.
\end{proof}

\begin{proposition}[Discontinuity of mapping teachers to optimal students]
\label{prop:counterexample}
Let $K_T=\{\vv_T(t):t\in[-\tfrac12,\tfrac12]\}\subset\mathcal V_T$.
There is no continuous map $T:K_T\to\mathcal V_S$ such that
$T(\vv_T(t))\in\operatorname{Opt}(\vv_T(t))$ for every $t$.
More strongly, for any continuous $T:K_T\to\mathcal V_S$ and any selection
$T^\star:K_T\to\mathcal V_S^0$ with
$T^\star(\vv_T(t))\in\operatorname{Opt}(\vv_T(t))$,
\begin{equation}
\label{eq:selection-gap}
 \sup_{t\in[-1/2,1/2]}
 \|f_{T(\vv_T(t))}-f_{T^\star(\vv_T(t))}\|_\infty
 \ge \frac12.
\end{equation}
\end{proposition}

\begin{proof}
The realization map $\vv_S\mapsto f_{\vv_S}|_I$ is continuous into $C(I)$
with the supremum norm.  Hence a continuous $T$ induces a continuous path
$\varphi(t)=f_{T(\vv_T(t))}|_I$.  If the supremum in
Equation~\ref{eq:selection-gap} were $\varepsilon<1/2$, then
Lemma~\ref{lem:best-hinge} and the limits $t\uparrow0$ and $t\downarrow0$
would give
\[
 \|\varphi(0)-u\|_\infty\le\varepsilon,
 \qquad
 \|\varphi(0)-v\|_\infty\le\varepsilon.
\]
The triangle inequality would imply
$1=\|u-v\|_\infty\le2\varepsilon<1$, a contradiction.
A continuous map selecting optimal students would have zero gap, so none
exists.
\end{proof}

\begin{remark}[The counterexample is in general position]
\label{rem:counterexample-general-position}
General position, as defined in Equation~\ref{eq:exclusion-set}, requires
distinct biases within each hidden layer.  Along the entire teacher path
$\vv_T(t)$, the hidden biases are $-10$ and $-9$, including at $t=0$.
The width-$1$ student satisfies this condition vacuously.  Hence
$K_T\subset\mathcal V_T\setminus E_{A_T}$ and $E_{A_S}=\varnothing$.
The approximation lower bound in Proposition~\ref{prop:counterexample}
therefore holds even in general position: the obstruction is the
discontinuous choice of an optimal student.
\end{remark}

\begin{remark}[Refinement from a student initialization]
The proposition concerns mapping each teacher to a globally optimal student.
A cross-architecture operator $F(\vv_T,\vv_0)$ instead refines a supplied
student initialization $\vv_0$, without requiring global optimality.
Appendix~\ref{sec:finite-step} proves that a fixed number of distillation
steps gives a continuous equivariant operator under the stated assumptions.
\end{remark}

\subsection{Finite-step distillation is a cross-architecture operator}
\label{sec:finite-step}

Recall from Section~\ref{sec:crossarch} that knowledge distillation trains
a student with parameters $\vv_S$ to match a teacher with parameters $\vv_T$
by minimizing
\[
L(\vv_T,\vv_S)
=\mathbb E_{\vx\sim\mathcal D}
 \left[\ell\!\left(f_{\vv_T}(\vx),f_{\vv_S}(\vx)\right)\right],
\]
where $\ell$ measures the discrepancy between their outputs
(Equation~\ref{eq:functional-loss}).
Starting from a student initialization $\vv_0$, this loss can be optimized
with standard first-order methods, including gradient descent
(GD)~\citep{bottou2018optimization}, stochastic gradient descent
(SGD)~\citep{bottou2012sgd}, momentum~\citep{polyak1964some},
RMSProp~\citep{tieleman2012rmsprop}, and Adam~\citep{kingma2015adam}.
These methods operate \emph{coordinatewise}: after computing the gradient,
they apply the same scalar update rule to each parameter, using its own
value and its current or past gradient components.  The gradient itself
may depend on all teacher and student parameters.

For full-batch optimization, set $L_j=L$ at every step.
For SGD, fix the sequence of sampled minibatches and let $L_j$ be the
average distillation loss over the minibatch at step $j$; the same
convention applies to minibatch versions of the other methods.
Assume each $L_j$ is differentiable in the student parameters and
$\nabla_{\vv_S}L_j(\vv_T,\vv_S)$ is jointly continuous in $(\vv_T,\vv_S)$.
Write $G_T$ and $G_S$ for the teacher and student permutation groups, and
$d_S=\dim\mathcal V_S$.

\begin{proposition}[Finite-step distillation]
\label{thm:finite-step}
Starting from $\vv^{(0)}=\vv_0$, suppose the optimizer has the coordinatewise
form
\begin{equation}
\label{eq:phi-k}
\begin{aligned}
 \vq^{(j)}&=\nabla_{\vv_S}L_j(\vv_T,\vv^{(j)}),\\
 v_i^{(j+1)}&=u_j\bigl((\vv_0)_i,q_i^{(0)},\ldots,q_i^{(j)}\bigr),
 \qquad 1\leq i\leq d_S,
\end{aligned}
\end{equation}
where each $u_j:\mathbb R^{j+2}\to\mathbb R$ is a fixed continuous function
shared across all coordinates.
Then, for every fixed integer $k\geq0$, the map
$\Phi_k(\vv_T,\vv_0):=\vv^{(k)}$ is continuous and satisfies
\[
 \Phi_k(g_T\!\cdot\vv_T,g_S\!\cdot\vv_0)
 =g_S\!\cdot\Phi_k(\vv_T,\vv_0)
 \qquad (g_T\in G_T,\;g_S\in G_S).
\]
Thus $\Phi_k$ is a continuous equivariant cross-architecture operator in the
sense of Definition~\ref{def:cross-architecture-operator}. Functionally
equivalent teachers produce identical trajectories from the same initialization.
\end{proposition}

\begin{proof}
Each $L_j$ depends only on the realized teacher and student functions, so
\begin{equation}
\label{eq:loss-symmetries}
 f_{\vv_T}=f_{\vv_T'}\Longrightarrow
 L_j(\vv_T,\cdot)=L_j(\vv_T',\cdot),
 \qquad
 L_j(\vv_T,P_g\vv_S)=L_j(\vv_T,\vv_S),
\end{equation}
where $P_g$ is the permutation matrix of $g\in G_S$.
Differentiating the second identity and using $P_gP_g^\top=I$ gives
\begin{equation}
\label{eq:gradient-equivariance}
 \nabla_{\vv_S}L_j(\vv_T,P_g\vv_S)
 =P_g\nabla_{\vv_S}L_j(\vv_T,\vv_S).
\end{equation}
Differentiating $L_j(\vv_T,\cdot)=L_j(\vv_T',\cdot)$ likewise shows that
functionally equivalent teachers give identical student gradients.

Continuity follows by induction: $\vv^{(0)}=\vv_0$ is continuous in
$(\vv_T,\vv_0)$, and the jointly continuous gradients and continuous $u_j$
make each subsequent iterate continuous by composition.

For symmetry, compare runs initialized at $(\vv_T,\vv_0)$ and
$(\vv_T',P_g\vv_0)$, where $f_{\vv_T'}=f_{\vv_T}$.
Denote their student parameters at step $r$ by $\vv^{(r)}$ and
$\widetilde{\vv}^{(r)}$, respectively.  Their student-loss gradient vectors are
\[
 \vq^{(r)}=\nabla_{\vv_S}L_r(\vv_T,\vv^{(r)}),
 \qquad
 \widetilde{\vq}^{(r)}
 =\nabla_{\vv_S}L_r(\vv_T',\widetilde{\vv}^{(r)}).
\]
Assume inductively that $\widetilde{\vv}^{(r)}=P_g\vv^{(r)}$ for
$0\leq r\leq j$.  Substituting this relation, replacing the teacher using
Equation~\ref{eq:loss-symmetries}, and applying the gradient transformation
in Equation~\ref{eq:gradient-equivariance} gives
\[
\begin{aligned}
 \widetilde{\vq}^{(r)}
 &=\nabla_{\vv_S}L_r(\vv_T',P_g\vv^{(r)})\\
 &=\nabla_{\vv_S}L_r(\vv_T,P_g\vv^{(r)})\\
 &=P_g\nabla_{\vv_S}L_r(\vv_T,\vv^{(r)})
  =P_g\vq^{(r)}
 \qquad(0\leq r\leq j).
\end{aligned}
\]
Writing $(P_gx)_i=x_{\pi_g(i)}$, the shared update rule in
Equation~\ref{eq:phi-k} therefore yields
\[
\begin{aligned}
 \widetilde v_i^{(j+1)}
 &=u_j\bigl((P_g\vv_0)_i,\widetilde q_i^{(0)},\ldots,
             \widetilde q_i^{(j)}\bigr)\\
 &=u_j\bigl((\vv_0)_{\pi_g(i)},q_{\pi_g(i)}^{(0)},\ldots,
             q_{\pi_g(i)}^{(j)}\bigr)
  =v_{\pi_g(i)}^{(j+1)}.
\end{aligned}
\]
Since $\widetilde{\vv}^{(0)}=P_g\vv_0$, induction gives
$\widetilde{\vv}^{(k)}=P_g\vv^{(k)}$ for every fixed $k$.
Taking $\vv_T'=g_T\!\cdot\vv_T$ proves the claimed equivariance, and taking
$g$ to be the identity proves equality of trajectories for functionally
equivalent teachers.
\end{proof}

\begin{lemma}[Standard gradient methods]
\label{lem:gd-optimizer}
Gradient descent, SGD, momentum, RMSProp, and Adam have the form in
\eqref{eq:phi-k}, with fixed scalar hyperparameter schedules, decay factors
in $[0,1)$, the usual zero initialization of running averages, and
$\epsilon>0$ in the denominators of RMSProp and Adam.
\end{lemma}

\begin{proof}
Gradient descent and SGD subtract a scalar multiple of the current gradient.
Unrolling momentum expresses each update as a weighted sum of the gradients
computed so far. Likewise, the running averages in RMSProp and Adam are
weighted sums of gradients or squared gradients, with coefficients independent
of the parameter coordinate. Their remaining operations are coordinatewise
square roots and division; Adam's bias corrections are fixed scalars at each
step. All these operations are continuous, since the squared-gradient averages
are nonnegative and $\epsilon>0$ keeps denominators positive.
Unrolling the parameter updates therefore gives a continuous function $u_j$
of the initial coordinate and its gradient history, identical across
coordinates, as required by~\eqref{eq:phi-k}.
\end{proof}

\subsection{Universal approximation of cross-architecture operators}
\label{sec:universality}

We now prove Theorem~\ref{thm:crossarch-universality}.  The proof has two
conceptual ingredients.  First, every continuous source-invariant,
target-equivariant map can be approximated by a finite sum of a
source-invariant scalar times a target-equivariant operator.  Second, CrossGMN
can realize such finite mixtures: it computes the two factors on their
respective parameter graphs, communicates the scalar through a cross edge,
and multiplies them on the target side.

\paragraph{General position and the fixed-architecture theorem.}
For an MLP architecture $A=(\vd,\sigma)$, let
\begin{equation}
\label{eq:exclusion-set}
 E_A=
 \left\{\vv:\text{two neurons in some hidden layer have equal biases}\right\}.
\end{equation}
The complement is the general-position set used by
\citet{dayan2026expressive}.
We use the following result.

\begin{theorem}[Fixed-architecture universality; \citet{dayan2026expressive}]
\label{thm:fixed-arch-universality}
On every compact $K\subset\mathcal V_A\setminus E_A$, GMNs uniformly approximate every continuous $G_A$-invariant functional and every continuous $G_A$-equivariant operator
$K\to\mathcal V_A$.
\end{theorem}

We may assume that each compact set contains every hidden-neuron permutation
of each of its elements.  To obtain this property, enlarge the sets to
\[
\widetilde K_\alpha
=\{g\!\cdot\vv:\vv\in K_\alpha,\ g\in G_\alpha\},
\qquad \alpha\in\{\mathrm{src},\mathrm{tgt}\}.
\]
Each enlarged set is a finite union of permuted copies of a compact set,
so it remains compact.  It also remains in general position, since
permuting neurons only reorders the distinct biases within each hidden layer.
Extend $F$ to the enlarged sets by
\[
\widetilde F(g_{\mathrm{src}}\!\cdot\vv_{\mathrm{src}},
             g_{\mathrm{tgt}}\!\cdot\vv_{\mathrm{init}})
=g_{\mathrm{tgt}}\!\cdot F(\vv_{\mathrm{src}},\vv_{\mathrm{init}}).
\]
The symmetry condition in Equation~\ref{eq:cross-equivariance} ensures that
different ways of writing the same permuted input pair give the same output.
The extension is
continuous on each permuted copy of
$K_{\mathrm{src}}\times K_{\mathrm{tgt}}$; since these are finitely many
closed sets and the definitions agree on their overlaps, it is continuous
on their union.  Uniform approximation on the enlarged sets implies the
same guarantee on the original sets.  We therefore use the enlarged sets
below, retaining the notation $K_{\mathrm{src}},K_{\mathrm{tgt}}$, and $F$.

\begin{lemma}[Invariant--equivariant finite mixture]
\label{lem:separable-mixture}
Let
$F:K_{\mathrm{src}}\times K_{\mathrm{tgt}}\to\mathcal V_{\mathrm{tgt}}$
be continuous, source invariant, and target equivariant.  For every
$\eta>0$, there are continuous source-invariant functions
$a_1,\ldots,a_M:K_{\mathrm{src}}\to[0,1]$ with
$\sum_i a_i\equiv1$ and continuous target-equivariant maps
$h_1,\ldots,h_M:K_{\mathrm{tgt}}\to\mathcal V_{\mathrm{tgt}}$ such that
\begin{equation}
\label{eq:mixture-bound}
 \sup_{(\vv_{\mathrm{src}},\vv_{\mathrm{init}})}
 \left\|
 F(\vv_{\mathrm{src}},\vv_{\mathrm{init}})
 -\sum_{i=1}^{M}a_i(\vv_{\mathrm{src}})h_i(\vv_{\mathrm{init}})
 \right\|<\eta.
\end{equation}
\end{lemma}

\begin{proof}
Let $Q=K_{\mathrm{src}}/G_{\mathrm{src}}$ be the source orbit space.  Because
the group is finite and acts by isometries,
\[
 d_Q([\vv],[\vv'])=
 \min_{g\in G_{\mathrm{src}}}\|g\!\cdot\vv-\vv'\|
\]
makes $Q$ a compact metric space.  Source invariance lets $F$ factor through a
continuous map
$\overline F:Q\times K_{\mathrm{tgt}}\to\mathcal V_{\mathrm{tgt}}$,
and every $\overline F(q,\cdot)$ remains target equivariant.  Uniform
continuity implies that
$q\mapsto\overline F(q,\cdot)$ is continuous in the supremum norm.  Choose a
finite cover $U_1,\ldots,U_M$ of $Q$ and points $q_i\in U_i$ such that
\[
 q\in U_i
 \quad\Longrightarrow\quad
 \|\overline F(q,\cdot)-\overline F(q_i,\cdot)\|_\infty<\eta.
\]
Let $\{\alpha_i\}$ be a continuous partition of unity subordinate to this
cover, set $a_i(\vv)=\alpha_i([\vv])$, and set
$h_i(\vv_{\mathrm{init}})=\overline F(q_i,\vv_{\mathrm{init}})$.  Then the
$a_i$ are invariant, the $h_i$ are equivariant, and the convex-combination
bound gives \eqref{eq:mixture-bound}.
\end{proof}

\begin{lemma}[CrossGMN realizes finite mixtures]
\label{lem:joint-mixture}
Let $K_{\mathrm{src}}$ and $K_{\mathrm{tgt}}$ be compact general-position
sets for fixed source and target MLP architectures.  For $i=1,\ldots,M$,
let $a_i:K_{\mathrm{src}}\to\mathbb R$ be a continuous source-invariant
functional, and let $h_i:K_{\mathrm{tgt}}\to\mathcal V_{\mathrm{tgt}}$ be
continuous and target equivariant.  For every $\varepsilon>0$, there exists a
$\text{CrossGMN}_\theta$ whose joint graph uses the positionwise boundary
joins of Section~\ref{sec:crossgmn} and the strided cross-edge policy of
Definition~\ref{def:strided-cross-edge-policy} such that
\begin{equation}
\label{eq:mixture-approx}
 \sup_{(\vv_{\mathrm{src}},\vv_{\mathrm{init}})
       \in K_{\mathrm{src}}\times K_{\mathrm{tgt}}}
 \left\|
 \text{CrossGMN}_\theta(\vv_{\mathrm{src}},\vv_{\mathrm{init}})
 -\sum_{i=1}^{M}a_i(\vv_{\mathrm{src}})h_i(\vv_{\mathrm{init}})
 \right\|<\varepsilon.
\end{equation}
The same conclusion holds if further symmetry-preserving cross edges are added.
\end{lemma}

\begin{proof}
\emph{Approximate the factors.}
Let $\ve_{\mathrm{out}}\in\mathcal V_{\mathrm{src}}$ have value $1$ in
the first output-bias coordinate and $0$ elsewhere, and define
$\Psi_i(\vv_{\mathrm{src}})=a_i(\vv_{\mathrm{src}})\ve_{\mathrm{out}}$.
This map is continuous and source equivariant because $a_i$ is invariant
and hidden-neuron permutations fix $\ve_{\mathrm{out}}$.
For any $\eta>0$, Theorem~\ref{thm:fixed-arch-universality} gives source and
target GMNs $S_i,H_i$ satisfying
\[
\sup_{K_{\mathrm{src}}}\|S_i-\Psi_i\|<\eta,
\qquad
\sup_{K_{\mathrm{tgt}}}\|H_i-h_i\|<\eta.
\]

\emph{Store and compute the factors in node and edge features.}
Let $x_v,x_{uv}$ be the original node and edge features, including the
parameter values and structural labels.  At step $j$, use
\[
z_v^{(j)}=\bigl(x_v,(w_{v,i}^{(j)})_{i=1}^M,c_v^{(j)}\bigr),
\qquad
z_{uv}^{(j)}=\bigl(x_{uv},(w_{uv,i}^{(j)})_{i=1}^M,c_{uv}^{(j)}\bigr).
\]
The block $w_{\cdot,i}$ stores the working state of $S_i$ on the source
side and of $H_i$ on the target side; add zero coordinates and identity
layers to match their dimensions and depths.  The vectors
$c_v,c_{uv}\in\mathbb R^M$ will store the source scalars and are initially
zero.  Keep $x_v,x_{uv}$ unchanged throughout.

Set $\chi_\alpha(v)=\mathbf 1_{\{v\in V^\alpha\}}$ and
$\chi_\alpha(u,v)=\chi_\alpha(u)\chi_\alpha(v)$, where
$\alpha\in\{\mathrm{src},\mathrm{tgt}\}$.
Initialize each working block with its side's GMN encoder applied to $x$.
For block $i$, suppress $i$ and the step index on its message, node-update,
and edge-update maps $\phi^\alpha,U^\alpha,E^\alpha$, and use
\[
\begin{aligned}
m_{uv,i}^{(j)}
&=\sum_\alpha\chi_\alpha(u,v)\,
 \phi^\alpha(w_{u,i}^{(j)},w_{v,i}^{(j)},w_{uv,i}^{(j)}),\\
w_{v,i}^{(j+1)}
&=\sum_\alpha\chi_\alpha(v)\,
 U^\alpha\!\left(w_{v,i}^{(j)},\sum_{u:u\to v}m_{uv,i}^{(j)}\right),\\
w_{uv,i}^{(j+1)}
&=\sum_\alpha\chi_\alpha(u,v)\,
 E^\alpha(w_{u,i}^{(j)},w_{v,i}^{(j)},w_{uv,i}^{(j)}).
\end{aligned}
\]
Thus each side performs its own GMN computation, and cross edges send zero
messages in this stage.  Apply the local readouts and replace the working
blocks by their scalar outputs.  Let $\lambda(p)$ be the node or edge
storing target parameter $p$, as in Appendix~\ref{app:mlp-graphs}, and let
$o_{\mathrm{src}}$ be the first source output node.  The resulting features
contain
\[
w_{o_{\mathrm{src}},i}
=\widehat a_i:=[S_i(\vv_{\mathrm{src}})]_{\mathrm{out}},
\qquad
w_{\lambda(p),i}
=[\widehat h_i]_p,\quad
\widehat h_i:=H_i(\vv_{\mathrm{init}}).
\]
Here $[\cdot]_{\mathrm{out}}$ selects the first output-bias coordinate.
Consequently $|\widehat a_i-a_i|<\eta$ and
$\|\widehat h_i-h_i\|<\eta$ uniformly on the corresponding input sets.

\emph{Copy the source scalars to the target features.}
Write $\widehat{\va}=(\widehat a_1,\ldots,\widehat a_M)$.
On the positionwise cross edge from $o_{\mathrm{src}}$ to the first target
output node $o_{\mathrm{tgt}}$, send the message
$(w_{o_{\mathrm{src}},i})_{i=1}^M=\widehat{\va}$ to the $c$ field.
Set all other messages in this step to zero, so that
\[
c_v^{(0)}=\mathbf 1_{\{v=o_{\mathrm{tgt}}\}}\widehat{\va}
\qquad(v\in V^{\mathrm{tgt}}).
\]
We now update only $c$, keeping the computed factors and original features
fixed.  Use the bidirectional internal message passing of
Appendix~\ref{app:core}.  Let $d(v)$ be the distance from $o_{\mathrm{tgt}}$
in the target graph, and let $N_-(v)$ consist of its target neighbors at
distance $d(v)-1$.  For $v\ne o_{\mathrm{tgt}}$, this set is nonempty.
Apply
\[
c_v^{(j+1)}=
\begin{cases}
c_v^{(j)},&v=o_{\mathrm{tgt}},\\[2pt]
\displaystyle\frac{1}{|N_-(v)|}\sum_{u\in N_-(v)}c_u^{(j)},
 &v\ne o_{\mathrm{tgt}}.
\end{cases}
\]
The required neighbor masks and normalization constants are determined by
the fixed MLP architecture, layer labels, and canonical output indices.
Induction on $d(v)$ gives $c_v^{(j)}=\widehat{\va}$ for $j\ge d(v)$.
After $D=\max_v d(v)$ steps, set $c_{uv}=c_v$ on each target edge.
Every target parameter location now has the features
\[
z_{\lambda(p)}
=\bigl(x_{\lambda(p)},([\widehat h_i]_p)_{i=1}^M,
                       (\widehat a_i)_{i=1}^M\bigr).
\]

\emph{Multiply the stored factors and read out the parameters.}
Multiplication is continuous on $[-R,R]^2$.  Hence, by the classical
universal approximation theorems~\citep{cybenko1989approximation,hornik1991approximation},
for every $R>0$ and $\kappa>0$ there is an MLP
$\mu_{R,\kappa}:\mathbb R^2\to\mathbb R$ such that
\begin{equation}
\label{eq:mlp-multiplication}
\sup_{(x,y)\in[-R,R]^2}
\bigl|\mu_{R,\kappa}(x,y)-xy\bigr|<\kappa.
\end{equation}
Let $B=\max_i\{\|a_i\|_\infty,\|h_i\|_\infty\}$ and
$P=\dim\mathcal V_{\mathrm{tgt}}$.  Choose
\[
0<\eta<\min\!\left\{1,\frac{\varepsilon}{3M(2B+1)}\right\},
\qquad
0<\kappa<\frac{\varepsilon}{3M\sqrt P},
\qquad R=B+1.
\]
Then $|\widehat a_i|,|[\widehat h_i]_p|<R$.
Using the values stored at $\lambda(p)$, define the local readout
\[
[\widehat F_\kappa]_p
=\sum_{i=1}^{M}\mu_{R,\kappa}(c_{\lambda(p),i},w_{\lambda(p),i}),
\qquad
[\bm\Delta]_p
=\frac{[\widehat F_\kappa]_p-[\vv_{\mathrm{init}}]_p}{s}.
\]
The initialization coordinate is available in $x_{\lambda(p)}$.
Thus the residual update of Definition~\ref{def:crossgmn} returns
$\widehat F_\kappa$.  In the Euclidean norm, uniformly on the input sets,
\[
\begin{aligned}
\left\|\widehat F_\kappa-\sum_i a_i h_i\right\|
&\le M\sqrt P\,\kappa
 +\sum_i\left(
 |\widehat a_i-a_i|\,\|h_i\|
 +|\widehat a_i|\,\|\widehat h_i-h_i\|\right)\\
&\le M\sqrt P\,\kappa+M\eta(2B+\eta)
 <\frac{2\varepsilon}{3}.
\end{aligned}
\]
All local rules above are continuous on compact neighborhoods of the
feature ranges, and their structural masks preserve the source and target
permutations.  Uniform MLP approximation of these rules and uniform
continuity of their finite composition give a choice of $\theta$ such that
\[
\sup_{K_{\mathrm{src}}\times K_{\mathrm{tgt}}}
\left\|\text{CrossGMN}_\theta-\widehat F_\kappa\right\|
<\frac{\varepsilon}{3}.
\]
The triangle inequality proves Equation~\ref{eq:mixture-approx}.
Additional cross edges are assigned zero messages in this construction.
\end{proof}

\begingroup
\newtheorem*{appuniversality}{Theorem~\ref{thm:crossarch-universality}}
\begin{appuniversality}[Universality of CrossGMN, restated]
Let $A_{\mathrm{src}}$ and $A_{\mathrm{tgt}}$ be fixed MLP architectures,
and let
\[
K_{\mathrm{src}}\subset
\mathcal V_{A_{\mathrm{src}}}\setminus E_{A_{\mathrm{src}}},
\qquad
K_{\mathrm{tgt}}\subset
\mathcal V_{A_{\mathrm{tgt}}}\setminus E_{A_{\mathrm{tgt}}}
\]
be compact sets, where $E_A$ is defined in Equation~\ref{eq:exclusion-set}.
Let $F:K_{\mathrm{src}}\times K_{\mathrm{tgt}}\to\mathcal V_{\mathrm{tgt}}$
be a continuous cross-architecture operator in the sense of
Definition~\ref{def:cross-architecture-operator}.
For joint graphs containing the positionwise boundary joins and strided
hidden-layer joins of Section~\ref{sec:crossgmn}, and every
$\varepsilon>0$, there exists a $\text{CrossGMN}_\theta$ such that
\[
\sup_{(\vv_{\mathrm{src}},\vv_{\mathrm{init}})
      \in K_{\mathrm{src}}\times K_{\mathrm{tgt}}}
\left\|\text{CrossGMN}_\theta(\vv_{\mathrm{src}},\vv_{\mathrm{init}})
      -F(\vv_{\mathrm{src}},\vv_{\mathrm{init}})\right\|
<\varepsilon.
\]
\end{appuniversality}
\endgroup

\paragraph{Proof of Theorem~\ref{thm:crossarch-universality}.}
Apply Lemma~\ref{lem:separable-mixture} with error $\varepsilon/2$ and write
$F_{\mathrm{mix}}=\sum_i a_i h_i$.  Lemma~\ref{lem:joint-mixture} gives a
CrossGMN within $\varepsilon/2$ of $F_{\mathrm{mix}}$.  The triangle
inequality with \eqref{eq:mixture-bound} gives the desired uniform
$\varepsilon$ approximation of $F$.

\subsection{One CrossGMN for finitely many source architectures}
\label{sec:any-dim}

The same argument applies when the source population contains finitely many
architectures and all map to one target architecture.  The only additional
requirement is that the source architecture be identifiable from the graph.

\begin{definition}[Architecture-identifiable family]
\label{def:architecture-identifiable}
A finite family
$A_{\mathrm{src}}^{(1)},\ldots,A_{\mathrm{src}}^{(m)}$ is
\emph{architecture identifiable} if a finite message-passing block can write
the one-hot architecture tag $e_r$ on every node of a joint graph whose source
has architecture $A_{\mathrm{src}}^{(r)}$.  An explicit family feature is
sufficient.  For finite MLP families, graph depth, layer types, and layer
cardinalities are also sufficient.
\end{definition}

\begin{theorem}[Finite heterogeneous source families]
\label{thm:multi-source-universality}
Let $A_{\mathrm{tgt}}$ be a fixed MLP architecture and let
$A_{\mathrm{src}}^{(1)},\ldots,A_{\mathrm{src}}^{(m)}$ be an
architecture-identifiable finite family of MLP architectures.  Let
\[
K_r\subset
\mathcal V_{A_{\mathrm{src}}^{(r)}}\setminus E_{A_{\mathrm{src}}^{(r)}}
\quad(r=1,\ldots,m),
\qquad
K_{\mathrm{tgt}}\subset
\mathcal V_{A_{\mathrm{tgt}}}\setminus E_{A_{\mathrm{tgt}}}
\]
be compact sets, with $E_A$ as in Equation~\ref{eq:exclusion-set}.
For each $r$, let
$F_r:K_r\times K_{\mathrm{tgt}}\to\mathcal V_{\mathrm{tgt}}$ be continuous,
source invariant, and target equivariant.
Then one CrossGMN, with one parameter set, can approximate all $F_r$
uniformly to any prescribed accuracy.
\end{theorem}

\begin{proof}
Fix $\varepsilon>0$.  For each $r$, Theorem~\ref{thm:crossarch-universality}
gives $\widehat F_r=\text{CrossGMN}_{\theta_r}$ with
\[
\sup_{K_r\times K_{\mathrm{tgt}}}\|\widehat F_r-F_r\|
<\frac{\varepsilon}{2}.
\]
Add zero hidden coordinates and identity layers so that these networks have
common hidden-state dimensions and depth.

Let $x_v$ and $x_{uv}$ denote the original node and edge features, including
their parameter values and structural labels.  On a joint graph with source
architecture $A_{\mathrm{src}}^{(r)}$, the identification block of
Definition~\ref{def:architecture-identifiable} writes the tag
$e_r\in\mathbb R^m$ on every source and target node; its $r$-th entry is $1$
and all other entries are $0$.  Copy this tag to each edge from either
endpoint.  At step $j$, use node and edge features
\[
z_v^{(j)}=(x_v,e_r,h_v^{(j)}),
\qquad
z_{uv}^{(j)}=(x_{uv},e_r,h_{uv}^{(j)}),
\]
where $h_v^{(j)}$ and $h_{uv}^{(j)}$ are the working hidden states.
Initialize these states to zero.  The original features and tags remain
unchanged throughout; the first encoding step fills the working states.

For each encoding, message, node-update, edge-update, or readout operation
$j$, let $\Phi_j^{(q)}$ be the corresponding rule of network $q$, with
inputs padded to common dimensions.  If $\tau$ is the stored tag and $z$
collects the operation's other inputs, use the shared rule
\[
\Phi_j(\tau,z)=\sum_{q=1}^{m}\tau_q\Phi_j^{(q)}(z).
\]
A message uses the tag at its receiving node; node and edge operations use
their own stored tags.  Since all tags equal $e_r$,
$\Phi_j(e_r,z)=\Phi_j^{(r)}(z)$ at every operation.  Induction over the
encoding, message-passing steps, and readout therefore shows that these
rules return exactly $\widehat F_r$ on $K_r\times K_{\mathrm{tgt}}$.

The rules $\Phi_j$ are continuous and can be uniformly approximated by MLPs
on compact neighborhoods of the local feature ranges
\citep{cybenko1989approximation,hornik1991approximation}.
There are finitely many rules, layers, and graph sizes, and each input set
is compact.  Uniform continuity of the resulting finite compositions
therefore lets us choose their approximation accuracies so that one
assembled CrossGMN satisfies
\[
\max_{1\le r\le m}\;
\sup_{K_r\times K_{\mathrm{tgt}}}
\|\text{CrossGMN}_\theta-\widehat F_r\|
<\frac{\varepsilon}{2}.
\]
The triangle inequality gives
\[
\max_{1\le r\le m}\;
\sup_{K_r\times K_{\mathrm{tgt}}}
\|\text{CrossGMN}_\theta-F_r\|<\varepsilon.
\]
All local rules are shared across architectures, so $\theta$ is a single
parameter set, as required.
\end{proof}

\subsection{Scope for CNN and Transformer weight spaces}
\label{sec:transformer-universality}

Proposition~\ref{app:prop-equivariance-families} establishes equivariance for
CNNs and Transformers directly from their implemented parameter graphs.  The
universality proof extends to these families as long as the fixed-architecture
invariant and equivariant universality for GMNs on their parameter graphs still hold.
The extension applies with respect to the neural graph automorphism groups defined in
Appendix~\ref{app:arch-symmetries}.

\providecommand{\norm}[1]{\left\|#1\right\|}

\section{Experimental Details}
\label{app:experiments}

Throughout,
\emph{CrossGMN} denotes the operator in
Equation~\ref{eq:crossgmn-update}, and the \emph{anchor} is the target
initialization $\vv_0$ that it refines.  As in Section~\ref{sec:exp_setup},
the source network is the teacher $\vv_T$ and the target network is the
student, whose predicted initialization is $\hat{\vv}_S$.

\subsection{CrossGMN implementation}
\label{app:core}

All experiments use the joint graph of the teacher $\vv_T$ and anchor
$\vv_0$.  An equivariant graph metanetwork processes this graph, and the
target-side readout predicts the residual in
Equation~\ref{eq:crossgmn-update}.

In the classifier experiments, node features concatenate a scalar, an $8$-dimensional
layer embedding, a $4$-dimensional source/target-role embedding, a
$4$-dimensional node-type embedding, and optional $4$-dimensional family and
activation embeddings.  Edge features concatenate the parameter value, the
layer embedding, a $4$-dimensional edge-role embedding, and an optional family
embedding.  In the INR experiments, node and edge inputs both have dimension
$1+4+4=9$: scalar value, layer embedding, and role embedding.  Parameter
values are not normalized.  The four edge roles are source-internal,
target-internal, source-to-target, and target-to-source.

\paragraph{Cross edges.}
Cross edges use the layerwise joins and matching notation of
Appendix~\ref{app:cross-policies}.  The formal joint graphs in
Appendices~\ref{app:mlp-graphs}--\ref{app:transformer-graphs}
use positionwise input/output joins.  The CNN and ViT classifier encoding
uses dense boundary joins, replacing $\operatorname{Pos}$ by
$\operatorname{Join}$ in Equation~\ref{eq:app-boundary-cross}; MLP boundaries
remain positionwise.  Both choices preserve the stated symmetries because
input and output nodes are fixed.

\paragraph{Backbone and readout.}
We use a bidirectional \textsc{ScaleGMN} backbone
\citep{kalogeropoulos2024scalegmn}, motivated by its better empirical
performance than plain GMN in most experiments.  Our universality results
are established for plain GMN.  The backbone uses skip connections, edge updates, and
dropout $0.1$ during operator training.  Message
and update functions are two-layer SiLU MLPs with layer normalization.  The
classifier families use scale-equivariant message passing; the SIREN families
use sign-equivariant message passing, except for the parameter-matched
permutation-equivariant ModelNet40 width+depth operator.  Table~\ref{tab:core-arch}
gives the configuration for each model family.

\begin{table}[h]
\centering
\small
\caption{CrossGMN configurations.  $d_v/d_e$ are node/edge input widths and
$|\theta|$ is the metanetwork parameter count; $s$ is the residual scale in
Equation~\ref{eq:crossgmn-update}.}
\label{tab:core-arch}
\resizebox{\linewidth}{!}{%
\begin{tabular}{llccccrc}
\toprule
Family / operator group & Symmetry & $d_{\rm hid}$ & Layers & Agg. & $d_v/d_e$ & $|\theta|$ & $s$ \\
\midrule
MNIST SIREN (all; depth and width+depth use $s=.01$) & sign & 128 & 4 & mean & $9/9$ & 2,647,895 & .005 \\
ModelNet40 depth $4\!\to\!2$ & sign & 192 & 4 & add & $9/9$ & 5,937,947 & .02 \\
ModelNet40 depth $4\!\to\!1$ & sign & 192 & 4 & add & $9/9$ & 5,937,907 & .02 \\
ModelNet40 width & sign & 192 & 4 & add & $9/9$ & 5,937,907 & .005 \\
ModelNet40 width+depth & permutation & 324 & 4 & add & $9/9$ & 6,010,354 & .01 \\
ViT depth & scale & 64 & 4 & mean & $25/17$ & 405,587 & .1 \\
ViT attention dimension, head count, head count+depth & scale & 128 & 6 & mean & $25/17$ & 2,438,163 & .1 \\
CNN width & scale & 64 & 6 & mean & $21/13$ & 620,607 & .4 \\
MLP width / depth / width+depth & scale & 128 & 8 & mean & $17/13$ & 3,293,663 & .1 \\
MLP heterogeneous architectures & scale & 128 & 4 & mean & $17/13$ & 1,579,487 & .1 \\
\bottomrule
\end{tabular}}
\end{table}

\paragraph{Operator objective.}
We train CrossGMN directly on teacher--student outputs, without an inner
distillation loop.  For a minibatch of $B$ teachers, write
\[
\hat{\vv}_{S,i}
=\text{CrossGMN}_\theta(\vv_{T,i},\vv_{0,i})
=\vv_{0,i}+s\,\bm\Delta_\theta(\vv_{T,i},\vv_{0,i}),
\qquad i=1,\ldots,B,
\]
where $\vv_{0,i}$ is the corresponding anchor.  For classifiers, a shared
image batch $X$ gives the objective
\begin{equation}
\label{eq:app-core-loss}
\mathcal L(\theta)
=\frac{1}{B|X|}\sum_{i=1}^{B}\sum_{\vx\in X}
\mathrm{KL}\!\left(
\operatorname{softmax}(f_{\vv_{T,i}}(\vx))
\,\middle\|\,
\operatorname{softmax}(f_{\hat{\vv}_{S,i}}(\vx))
\right).
\end{equation}
The temperature is $1$; operator training uses neither a cross-entropy
term nor logit clipping.

For INRs, we minimize the MSE between teacher and predicted-student
outputs.  Let $\mathcal G\subset\mathbb R^2$ be the full $28\times28$ pixel
grid and $\mathcal P_i\subset\mathbb R^3$ the fixed set of uniformly sampled
query points for ModelNet40 teacher $i$.  The respective objectives are
\begin{align}
\mathcal L_{\mathrm{render}}(\theta)
&=\frac{1}{B|\mathcal G|}
  \sum_{i=1}^{B}\sum_{\vx\in\mathcal G}
  \left|f_{\hat{\vv}_{S,i}}(\vx)-f_{\vv_{T,i}}(\vx)\right|^2,
\label{eq:app-inr-render-loss}\\
\mathcal L_{\mathrm{field}}(\theta)
&=\frac1B\sum_{i=1}^{B}\frac{1}{|\mathcal P_i|}
  \sum_{\vx\in\mathcal P_i}
  \left|f_{\hat{\vv}_{S,i}}(\vx)-f_{\vv_{T,i}}(\vx)\right|^2.
\label{eq:app-inr-field-loss}
\end{align}
The scalar outputs represent pixel intensities for MNIST and distance-field
values for ModelNet40.  Table~\ref{tab:core-train} gives the training
hyperparameters and number of query points.

\begin{table}[h]
\centering
\small
\caption{Operator training.}
\label{tab:core-train}
\resizebox{\linewidth}{!}{%
\begin{tabular}{lccccc}
\toprule
 & ViT & CNN & MLP & MNIST INR & ModelNet40 INR \\
\midrule
Objective & Eq.~\ref{eq:app-core-loss} & Eq.~\ref{eq:app-core-loss} & Eq.~\ref{eq:app-core-loss} & Eq.~\ref{eq:app-inr-render-loss} & Eq.~\ref{eq:app-inr-field-loss} \\
Optimizer / decay & AdamW / $10^{-3}$ & AdamW / $10^{-3}$ & AdamW / $10^{-3}$ & AdamW / $10^{-3}$ & AdamW / $10^{-3}$ \\
Learning rate & $3\!\cdot\!10^{-4}$ or $10^{-3}$ & $10^{-4}$ & $10^{-4}$ / $10^{-3}$ & $5\!\cdot\!10^{-4}$--$2\!\cdot\!10^{-3}$ & $3$--$5\!\cdot\!10^{-4}$ \\
Warmup (steps) & none & 200 & 200 & none & none \\
Functional inputs & 128 images & 128 images & 128 images & 784 coordinates & 4096 or 2048 points \\
\bottomrule
\end{tabular}}
\end{table}
All runs use global gradient clipping at $1$.

\subsection{Evaluation protocol}
\label{app:protocol}

\paragraph{Teacher, anchor, and predicted student.}
We follow the notation of Section~\ref{sec:exp_setup}.  For each held-out
teacher $\vv_T$, we construct the anchor $\vv_0$ using the corresponding
method in Table~\ref{tab:anchor-grid} and predict
$\hat{\vv}_S=\text{CrossGMN}_\theta(\vv_T,\vv_0)$.
Let $\vv_S^a(k)$ denote the student parameters after $k$ distillation steps,
where $a\in\{\mathrm{anchor},\mathrm{CrossGMN}\}$ identifies the initialization:
\[
\vv_S^{\mathrm{anchor}}(0)=\vv_0,
\qquad
\vv_S^{\mathrm{CrossGMN}}(0)=\hat{\vv}_S.
\]
Both students are distilled toward the same teacher with the same optimizer,
minibatches, data order, and evaluation steps.  The comparison
therefore measures the benefit of refining the anchor with CrossGMN.
Additional structured-pruning baselines and random initialization appear in
the reach--budget plots.

\begin{table}[h]
\centering
\small
\caption{Anchor initializations used for the distillation comparisons.}
\label{tab:anchor-grid}
\begin{tabular}{@{}p{.18\linewidth}p{.26\linewidth}p{.48\linewidth}@{}}
\toprule
Family & Reduction axis & Anchor \\
\midrule
MNIST SIREN & width / depth / width+depth & magnitude pruning / layer fold / fold, then prune \\
ModelNet40 SIREN & width / depth / width+depth & magnitude pruning / strided layer transfer / strided prune \\
CIFAR-10 MLP & width / depth or width+depth & magnitude pruning / layer fold (scale $0.1$) \\
Heterogeneous MLPs & all & layer fold (scale $0.1$) \\
CIFAR-10 ViT & attention dimension / head count / depth / head count+depth & top-$k$ attention dimensions / head drop / strided block transfer / strided blocks + head drop \\
CNN & channel width & magnitude pruning \\
\bottomrule
\end{tabular}
\end{table}

\paragraph{Student initializations.}
\label{app:student-init}
We define the structured initializations used as anchors and comparison
baselines below; Table~\ref{tab:anchor-grid} identifies the anchor for each
setting.  Write the teacher's affine layers as
$(\mW_\ell,\vb_\ell)_{\ell=1}^{L_T}$, with hidden widths $d_\ell$, and let
$L_S$ be the student's number of affine layers.  Retained unit indices are
kept in increasing order.

\paragraph{Width reduction.}
When $L_S=L_T$, selecting $m_\ell$ units in each hidden layer gives index
sets $S_\ell\subseteq\{1,\ldots,d_\ell\}$ with $|S_\ell|=m_\ell$.
With $S_0$ and $S_{L_T}$ containing all input and output indices, the
student parameters are
\[
\mW_\ell^0=\mW_\ell[S_\ell,S_{\ell-1}],
\qquad
\vb_\ell^0=\vb_\ell[S_\ell].
\]
Magnitude-based pruning~\citep{xu2024initializing,li2017pruning} and
activation-based pruning~\citep{hu2016network} motivate the following
variants.  For each hidden layer, our implementations keep the $m_\ell$
units with the largest scores:
\begin{align}
s_{\ell i}^{\mathrm{magnitude}}
 &=\textstyle\sum_j |(\mW_\ell)_{ij}|+
                  \sum_j |(\mW_{\ell+1})_{ji}|,\\
s_{\ell i}^{\ell_1\text{-outgoing}}
 &=\textstyle\sum_j |(\mW_{\ell+1})_{ji}|,\\
s_{\ell i}^{\mathrm{magnitude\text{-}sum}}
 &=\textstyle\sum_j |(\mW_\ell)_{ij}|,\\
s_{\ell i}^{\mathrm{activation}}
 &=\frac{1}{n_{\mathrm{cal}}}\sum_{r=1}^{n_{\mathrm{cal}}}
   \left|\left[\sigma(\mW_\ell\vx_{\ell-1,r}+\vb_\ell)\right]_i\right|.
\end{align}
The magnitude anchor uses a two-sided $\ell_1$ score, summing incoming and
outgoing weight magnitudes.  Here $\vx_{\ell-1,r}$ is the input to layer $\ell$ for calibration example
$r$, and $\sigma$ is its activation.  Two baselines use no score:
\emph{uniform} selection retains
\[
S_\ell=\left\{1+\operatorname{round}\!\left(
\frac{r(d_\ell-1)}{m_\ell-1}\right):r=0,\ldots,m_\ell-1\right\}
\]
for $m_\ell>1$ (and $S_\ell=\{1\}$ for $m_\ell=1$);
\emph{random-neuron} selection samples $m_\ell$ distinct indices using a
fixed seed per teacher.

\paragraph{CNN magnitude pruning.}
For the CNN kernels $K_\ell$ of Appendix~\ref{app:cnn-graphs}, the anchor
uses the convolutional analogue of the MLP magnitude score.  For output
channel $c$ of layer $\ell$, followed by another convolution, this is
\[
s_{\ell c}^{\mathrm{magnitude}}
=\sum_{i,r,t}|(K_\ell)_{cirt}|
 +\sum_{o,r,t}|(K_{\ell+1})_{ocrt}|,
\]
where $i,o$ index channels and $r,t$ index spatial kernel entries.
For the final convolution, the outgoing term uses the first dense layer's
weights associated with that channel.  We retain the highest-scoring
channels in increasing index order and slice biases and adjacent weights
consistently.  Thus the anchor uses structured magnitude pruning with a
two-sided $\ell_1$ score; the incoming term alone is the filter criterion of
\citet{li2017pruning}.  For ViTs, structured selection acts on heads or
attention coordinates and slices the corresponding projection blocks.

\paragraph{Structured OBS/OBC.}
This baseline removes units sequentially while correcting the surviving
weights using a quadratic layer-reconstruction objective
\citep{hassibi1992obs,frantar2022obc}.  For hidden layer $\ell$, let
$\mA\in\mathbb R^{n_{\mathrm{cal}}\times d_\ell}$ contain its calibration
activations as rows, and define
\[
\mH_0=\frac{\mA^\top\mA}{n_{\mathrm{cal}}},
\qquad \bar h=\frac{\operatorname{tr}(\mH_0)}{d_\ell},
\qquad \mH=\mH_0+10^{-2}\bar h\mI.
\]
Starting from $\mW=\mW_{\ell+1}$ and $\mK=\mH^{-1}$, each iteration selects
and compensates for a removed unit by
\begin{equation}
p=\argmin_i\frac{\|\mW[:,i]\|_2^2}{\mK_{ii}},
\qquad
\mW\leftarrow\mW-\frac{\mW[:,p]\mK[p,:]}{\mK_{pp}}.
\end{equation}
Delete column $p$ of $\mW$ and the corresponding row and bias entry in the
preceding affine map, and update the inverse on the remaining indices:
\[
\mK\leftarrow\left(
\mK-\frac{\mK[:,p]\mK[p,:]}{\mK_{pp}}\right)_{-p,-p}.
\]
Repeat until $m_\ell$ units remain.

\paragraph{Depth and joint reduction.}
For $2\le L_S<L_T$, \emph{strided layer transfer} copies student layer $j$
from teacher layer
\begin{equation}
\pi(j)=1+\operatorname{round}\!\left(
\frac{(j-1)(L_T-1)}{L_S-1}\right),
\qquad j=1,\ldots,L_S.
\end{equation}
\emph{Layer fold} instead replaces each contiguous group
$\ell_1,\ldots,\ell_2$ by the affine composition obtained by omitting its
intermediate nonlinearities:
\begin{equation}
\mW^0=c\,\mW_{\ell_2}\cdots\mW_{\ell_1},
\qquad
\vb^0=c\left(\vb_{\ell_2}+
\sum_{r=\ell_1}^{\ell_2-1}
\mW_{\ell_2}\cdots\mW_{r+1}\vb_r\right).
\end{equation}
At $c=1$ these are exactly the parameters of that affine composition;
the scale controls its magnitude.  Folded classifier maps use $c=0.1$,
whereas INRs use $c=1$.
\emph{Keep-first} copies the first $L_S-1$ teacher layers and initializes a
new output layer; \emph{keep-last} copies the last $L_S-1$ layers and
initializes a new input layer; \emph{keep-first+last} retains layers at both
ends and drops intermediate ones.  For width+depth reduction,
\emph{fold-then-prune} applies magnitude pruning after folding, and
\emph{strided-prune} applies it after strided layer selection.
Fresh random initialization uses the corresponding model family's
initialization rule.  Family-specific choices are given below.

\paragraph{Quality and target.}
Write $q_{\vv_T}^a(k)$ for the quality of $\vv_S^a(k)$, with larger values
indicating better agreement with the teacher.  For classifiers, this is the
fraction of test inputs on which teacher and student predict the same
label, measured on the full test set for ViTs and CNNs and a fixed
$1{,}000$-image subset for MLPs.  For INRs, it is the negative MSE between
student and teacher renders or fields.  Both initializations are evaluated
on the step grid $\mathcal G=\{0,10,20,\ldots\}$, without interpolation.

For each teacher, we determine the anchor's plateau step $p_{\vv_T}$ using
the improvement threshold $\epsilon=10^{-4}$ and patience $P=50$
evaluations from Section~\ref{sec:exp_setup}.  Its target quality is
\[
q_{\vv_T}^{\star}
=\max_{\substack{k\in\mathcal G\\k\le p_{\vv_T}}}
 q_{\vv_T}^{\mathrm{anchor}}(k).
\]
Thus each predicted student is evaluated against the quality attained by
its own teacher's anchor.

\paragraph{Budget and per-teacher speedup.}
Let $\mathcal T$ be the set of $N$ held-out teachers in a compression
setting.  The setting's budget is the empirical $95$th percentile of the
anchor plateau steps, rounded up to the next evaluation step:
\begin{equation}
\label{eq:budget-B}
B=10\left\lceil
\frac{Q_{0.95}(\{p_{\vv_T}:\vv_T\in\mathcal T\})}{10}
\right\rceil.
\end{equation}
The CrossGMN distillation budget is $H=1.5B$.
For $a\in\{\mathrm{anchor},\mathrm{CrossGMN}\}$, define the first recorded
step reaching the target by
\[
k_{\vv_T}^{a}
=\min\left\{k\in\mathcal G:
 k\le h_{\vv_T}^{a},\
 q_{\vv_T}^{a}(k)\ge q_{\vv_T}^{\star}\right\},
\qquad
h_{\vv_T}^{\mathrm{anchor}}=p_{\vv_T},\quad
h_{\vv_T}^{\mathrm{CrossGMN}}=H,
\]
where the minimum is over recorded evaluations and
$\min\varnothing=+\infty$.  The anchor reaches its target by
construction.  Following the main paper, the per-teacher speedup is
\[
S_{\vv_T}
=\frac{k_{\vv_T}^{\mathrm{anchor}}}
       {k_{\vv_T}^{\mathrm{CrossGMN}}}.
\]
If no recorded CrossGMN evaluation reaches the target by $H$, we count a
failure and set $S_{\vv_T}=0$.

\paragraph{Reported metrics.}
We report the median and interquartile range of
$\{S_{\vv_T}:\vv_T\in\mathcal T\}$, together with the acceleration and
failure rates defined in the main paper:
\[
r_{\mathrm{acc}}
=\frac1N\sum_{\vv_T\in\mathcal T}\mathbf1[S_{\vv_T}>1],
\qquad
r_{\mathrm{fail}}
=\frac1N\sum_{\vv_T\in\mathcal T}\mathbf1[S_{\vv_T}=0].
\]
All $N$ teachers, including failures, contribute to these statistics;
tables express the rates as percentages.  The additional $>2\times$
column in Table~\ref{tab:full} uses the same definition with threshold $2$.
The reach--budget plots instead track fidelity to the teacher and
performance retention against teacher-relative thresholds, as described in
Section~\ref{sec:exp_setup} and Appendix~\ref{app:reach}.

\paragraph{Wall-clock speedup.}
We include the time needed to construct each student initialization.
Let $c_{\mathrm{anchor}}$ be the anchor-construction time,
$c_{\mathrm{CrossGMN}}$ the complete CrossGMN initialization time
(including anchor construction, graph construction, and the forward pass),
and $d_{\mathrm{anchor}},d_{\mathrm{CrossGMN}}$ the measured distillation
time per step.  The per-teacher wall-clock speedup is
\begin{equation}
\label{eq:wall-speedup}
S_{\vv_T}^{\mathrm{wall}}
=
\frac{c_{\mathrm{anchor}}
      +d_{\mathrm{anchor}}k_{\vv_T}^{\mathrm{anchor}}}
     {c_{\mathrm{CrossGMN}}
      +d_{\mathrm{CrossGMN}}k_{\vv_T}^{\mathrm{CrossGMN}}}.
\end{equation}
Table~\ref{tab:full} reports the median over all teachers, assigning
wall-clock speedup $0$ to failures.  All timing constants were measured with
device synchronization and batch size one on NVIDIA
A100-SXM4-40GB (A100-SXM4-80GB for ViT).
The ModelNet40 width+depth setting has no recorded runtime profile.

\begin{table}[h]
\centering
\small
\caption{Evaluation-time distillation.  Classifiers use pure KL at temperature
$1$ and no learning-rate schedule; all evaluation loops record
every $10$ steps.}
\label{tab:proto-distill}
\resizebox{\linewidth}{!}{%
\begin{tabular}{lccccc}
\toprule
 & ViT & CNN & MLP & MNIST INR & ModelNet40 INR \\
\midrule
Optimizer & AdamW & AdamW & AdamW & Adam & Adam \\
Learning rate & $10^{-3}$ & $10^{-3}$ & $10^{-3}$ & $10^{-2}$ & $10^{-2}$ \\
Batch & 256 images & 256 images & 256 images & full $28\times28$ grid & 4096 points \\
Augmentation & none & none & crop + flip & -- & -- \\
Cohort & 1,000 & 400 & 1,000 (500 OOD) & 1,000 & 1,000 \\
\bottomrule
\end{tabular}}
\end{table}

\subsection{MNIST and FashionMNIST INR compression}
\label{app:inr}

\paragraph{Population and splits.}
We use the pre-fit MNIST SIREN population of
\citet{navon2023equivariant}: one INR per $28\times28$ image, with teacher
architecture $[2,32,32,1]$ ($1{,}185$ parameters), two sine hidden layers
with $\omega_0=30$, and a linear output.  Of $70{,}000$ INRs, $55{,}000$ are
used for operator training, $5{,}000$ for validation, and a fixed
$1{,}000$-INR subset of the test split for every reported evaluation.  The
FashionMNIST transfer population has the same architecture and
$55{,}000/5{,}000/10{,}000$ split.  The MNIST-trained operators are applied to
its $1{,}000$-INR test cohort without retraining.

\begin{table}[h]
\centering
\small
\caption{MNIST SIREN configurations.  Compression is the teacher-to-student
parameter-count ratio; $s$ is the CrossGMN residual scale.}
\label{tab:inr-arch}
\resizebox{\linewidth}{!}{%
\begin{tabular}{llrrlcc}
\toprule
Reduction axis & Student & Params & Compression & Anchor & Operator lr & $s$ \\
\midrule
Width & $[2,16,16,1]$ & 337 & $3.52\times$ & magnitude pruning & $10^{-3}$ & .005 \\
Width & $[2,8,8,1]$ & 105 & $11.29\times$ & magnitude pruning & $2\!\cdot\!10^{-3}$ & .005 \\
Width & $[2,4,4,1]$ & 37 & $32.03\times$ & magnitude pruning & $10^{-3}$ & .005 \\
Depth & $[2,32,1]$ & 129 & $9.19\times$ & layer fold & $5\!\cdot\!10^{-4}$ & .01 \\
Width+depth & $[2,16,1]$ & 65 & $18.23\times$ & fold, then prune & $5\!\cdot\!10^{-4}$ & .01 \\
\bottomrule
\end{tabular}}
\end{table}

\paragraph{Anchors and comparator initializations.}
Table~\ref{tab:inr-arch} specifies the anchors; their formulas and the
comparison baselines are defined in Appendix~\ref{app:student-init}.
For MNIST depth reduction, layer folding retains the first affine map and
composes the final two, omitting the intermediate sine activation.
The reach--budget plots compare CrossGMN with these anchors, the width or
depth baselines appropriate to each setting, and fresh SIREN initialization,
all distilled with the same optimizer and minibatches.  Width reductions
use strided cross edges; depth and width+depth reductions use all-to-all
cross edges.

\paragraph{Operator training and evaluation.}
The five operators use the MNIST row of Tables~\ref{tab:core-arch} and
\ref{tab:core-train}.  One epoch contains $1{,}719$ optimizer steps over the
$55{,}000$ training teachers; validation render MSE selects the checkpoint.
The selected checkpoints are from epochs $29,142,30,12,$ and $40$ in the
order of Table~\ref{tab:inr-arch}.  Evaluation distills the complete
$1{,}000$-teacher cohort as a batch of independent student parameter vectors and records per-INR
render MSE every ten steps.

\subsection{ModelNet40 3-D INR compression}
\label{app:modelnet}

\paragraph{Population and splits.}
We use the pre-fit SIREN distance-field population of
\citet{shamsian2024augmentations}, with teacher architecture
$[3,48,48,48,48,1]$ ($7{,}297$ parameters) and $\omega_0=30$.  The dataset
contains $12{,}311$ objects, each fit ten times; we use view $0$ and the
object-disjoint $7{,}874/1{,}969/2{,}468$ train/validation/test split.

\begin{table}[h]
\centering
\small
\caption{ModelNet40 configurations.  Columns follow Table~\ref{tab:inr-arch}.}
\label{tab:mn-arch}
\resizebox{\linewidth}{!}{%
\begin{tabular}{llrrlcc}
\toprule
Reduction axis & Student & Params & Compression & Anchor & Operator lr & $s$ \\
\midrule
Width & $[3,24,24,24,24,1]$ & 1,921 & $3.80\times$ & magnitude pruning & $3\!\cdot\!10^{-4}$ & $.005$ \\
Depth & $[3,48,48,1]$ & 2,593 & $2.81\times$ & strided layer transfer & $3\!\cdot\!10^{-4}$ & $.02$ \\
Depth & $[3,48,1]$ & 241 & $30.28\times$ & strided layer transfer & $5\!\cdot\!10^{-4}$ & $.02$ \\
Width+depth & $[3,24,24,1]$ & 721 & $10.12\times$ & strided prune & $3\!\cdot\!10^{-4}$ & $.01$ \\
\bottomrule
\end{tabular}}
\end{table}

\paragraph{Anchors, training, and evaluation.}
We use magnitude pruning for width reduction, strided layer transfer for
depth reduction, and strided-prune for width+depth reduction, as defined in
Appendix~\ref{app:student-init}.  Three operators use
the sign-equivariant $d_{\rm hid}=192$ backbone.  The width+depth operator uses
the parameter-matched permutation-equivariant $d_{\rm hid}=324$ backbone.  Each
is trained for $30$ epochs with a fixed set of uniformly sampled points and the
final epoch is evaluated.

All settings are evaluated on $4{,}096$ points.

\subsection{CIFAR-10 MLP compression}
\label{app:mlp}

\paragraph{Teacher populations.}
For fixed-architecture compression, we construct a model zoo of $6{,}100$ ReLU MLP teachers of
architecture $[3072,64,64,64,64,10]$ ($209{,}802$ parameters), trained for
$100$ epochs with varied learning rates, weight decay, label smoothing,
dropout, and seeds.  The split contains $5{,}000$ training, $100$
validation, and $1{,}000$ test teachers.

For heterogeneous-architecture compression, we construct a second ReLU MLP
model zoo with four training architectures:
$[3072,16,10]$, $[3072,16,16,10]$, $[3072,32,10]$, and
$[3072,32,32,10]$, all mapped to the common student $[3072,8,10]$ as in
Section~\ref{sec:multi-architecture}.  We hold out the wider architecture
$[3072,64,10]$ and the deeper architecture $[3072,16,16,16,16,10]$ to test
generalization to unseen source architectures.  The respective
splits contain $5{,}000$ training, $100$ validation, $1{,}000$ test teachers
from the training architectures, and $500$ teachers from unseen
architectures, balanced across the architectures in each split.
These model zoos and their fixed splits can also serve as benchmarks for
other weight-space methods, supporting evaluation within an architecture
and generalization to unseen architectures.

\begin{table}[h]
\centering
\small
\caption{MLP compression configurations.}
\label{tab:mlp-settings}
\resizebox{\linewidth}{!}{%
\begin{tabular}{llrrlcc}
\toprule
Reduction axis & Student & Params & Compression & Anchor & Operator lr & $s$ \\
\midrule
Depth & $[3072,64,64,10]$ & 201,482 & $1.04\times$ & layer fold (scale $0.1$) & $10^{-4}$ & $.1$ \\
Width & $[3072,32,32,32,32,10]$ & 101,834 & $2.06\times$ & magnitude pruning & $10^{-4}$ & $.1$ \\
Width+depth & $[3072,32,32,10]$ & 99,722 & $2.10\times$ & layer fold (scale $0.1$) & $10^{-4}$ & $.1$ \\
Heterogeneous architectures & $[3072,8,10]$ & 24,674 & $2.00$--$4.04\times$ & layer fold (scale $0.1$) & $10^{-3}$ & $.1$ \\
\bottomrule
\end{tabular}}
\end{table}

\paragraph{Anchors.}
We use the magnitude-pruning and layer-fold initializations defined in
Appendix~\ref{app:student-init}, with the choices in
Table~\ref{tab:mlp-settings}.  Layer folding partitions the teacher's affine
layers into contiguous groups, composes each group, and scales folded
interior maps by $0.1$.

\paragraph{Operator training and evaluation.}
The fixed-architecture operators use eight message-passing layers and
learning rate $10^{-4}$.  The heterogeneous-architecture operator uses four
layers and learning rate $10^{-3}$.
The test cohorts contain $1{,}000$ teachers from training architectures or
$500$ teachers from unseen architectures.

\subsection{CIFAR-10 Vision Transformer compression}
\label{app:vit}

\paragraph{Teacher population.}
We construct a model zoo of $6{,}100$ ViTs with patch size $4$, residual width $48$, four
blocks, two heads of dimension $24$, FFN width $192$, and a ten-class head
($115{,}930$ parameters).  Teachers vary in optimizer hyperparameters and
seed. The splits
contain $5{,}000$ training, $100$ validation, and $1{,}000$ test
teachers.  Like the MLP zoos, this population and its fixed splits
can serve as a benchmark for learning from model parameters, extending
evaluation to the Transformer family.
Operator configurations and checkpoints are chosen using only the
validation split.

\begin{table}[h]
\centering
\small
\caption{ViT compression configurations.  All four settings use the strided
cross-edge policy of Definition~\ref{def:strided-cross-edge-policy}; when the
teacher and student block counts agree it reduces to shared-index matching.
Here $d$ is the residual width, $L$ the number of blocks, $H$ the number of
heads, and $d_a$ the total attention dimension across heads.}
\label{tab:vit-settings}
\resizebox{\linewidth}{!}{%
\begin{tabular}{llrrlcc}
\toprule
Reduction axis & Student & Params & Compression & Anchor & Operator lr & $s$ \\
\midrule
Attention dimension & $d=48,L=4,H=2,d_a=24$ & 97,210 & $1.19\times$ & top-$k$ attention dimensions & $3\!\cdot\!10^{-4}$ & $.1$ \\
Head count & $d=48,L=4,H=1,d_a=24$ & 97,210 & $1.19\times$ & head drop & $10^{-3}$ & $.1$ \\
Depth & $d=48,L=2,H=2,d_a=48$ & 59,386 & $1.95\times$ & strided block transfer & $3\!\cdot\!10^{-4}$ & $.1$ \\
Head count+depth & $d=48,L=2,H=1,d_a=24$ & 50,026 & $2.32\times$ & strided blocks + head drop & $10^{-3}$ & $.1$ \\
\bottomrule
\end{tabular}}
\end{table}

\paragraph{Anchors, training, and evaluation.}
Attention-dimension and head-count anchors use two-sided $\ell_1$
magnitude pruning.  Each attention coordinate is scored by summing the absolute weights in its
value-projection row and output-projection column.  Attention-dimension
reduction retains the twelve highest-scoring coordinates in each head;
head-count reduction retains the head with the largest sum of coordinate
scores.  The retained indices are applied to the query, key, and value
rows and output columns, preserving their original order.  Depth reduction
uses strided block selection, retaining the first and third teacher blocks;
head count+depth reduction first selects these blocks and then prunes their
heads.
Parameters outside the reduced modules are copied unchanged.

Each operator is trained for $10{,}000$ steps at three candidate learning
rates; mean validation agreement selects the run and checkpoint.  The
selected checkpoints are from steps $10{,}000$, $9{,}500$, $9{,}000$, and
$10{,}000$, in the order of Table~\ref{tab:vit-settings}.

\subsection{MNIST CNN compression}
\label{app:cnn}

The public zoo of \citet{schurholtzoo} contains $2{,}560$ classifiers with
three convolutional stages of widths $[8,6,4]$, kernel sizes $5,5,2$, pooling
after the first two stages, a $20$-unit fully connected layer, and a ten-class
head ($2{,}464$ parameters).  The population spans tanh, ReLU, sigmoid, and
GELU models; the activation-type embedding is therefore enabled.

We reduce only convolutional width, $[8,6,4]\to[4,3,2]$, leaving the head
unchanged and retaining $41.5\%$ of the parameters.  The anchor uses the
two-sided $\ell_1$ magnitude-pruning criterion in Appendix~\ref{app:student-init}.
The operator uses
the CNN row of Tables~\ref{tab:core-arch} and \ref{tab:core-train}, with
residual scale $s=.4$, and is trained for $100{,}000$ steps on $1{,}407$
teachers.  Evaluation uses a fixed $400$-teacher cohort.

\subsection{Ablations}
\label{app:ablations-main}

All rows in Table~\ref{tab:ablations-main} use the MNIST INR reduction
$[2,32,32,1]\to[2,8,8,1]$ and the same $1{,}000$ held-out teachers as the
corresponding main result.  Each row changes one choice from the reference CrossGMN configuration
and follows the per-teacher speedup protocol in Appendix~\ref{app:protocol}.

\begin{resulttable}[t]{tab:ablations-main}{\textbf{CrossGMN ablations on MNIST INR compression.} Median speedup and IQR relative to the corresponding anchor, including failures ($S_{\vv_T}=0$).}
\shrinktofit{%
\begin{tabular}{llrrr}
\toprule
Ablation & Variant & \multicolumn{1}{l}{Speedup} & $>1\!\times$ (\%) & Failed (\%) \\
\midrule
Communication & Strided & $5.20\!\times_{[1.1,12.5]}$ & 76 & 18 \\
& No cross edges & $3.18\!\times_{[0.4,8.6]}$ & 67 & 24 \\
& All-to-all & $5.13\!\times_{[0.9,11.6]}$ & 75 & 18 \\
& FiLM on pooled teacher \citep{perez2018film} & $3.22\!\times_{[0.4,9.1]}$ & 68 & 24 \\
\midrule
Initialization & Structured pruning & $5.20\!\times_{[1.1,12.5]}$ & 76 & 18 \\
& Random initialization & $3.76\!\times_{[0.5,8.5]}$ & 70 & 23 \\
\midrule
Supervision & Teacher matching & $5.20\!\times_{[1.1,12.5]}$ & 76 & 18 \\
& Student after 50 steps & $1.15\!\times_{[0.0,4.0]}$ & 52 & 35 \\
& Student after 250 steps & $2.35\!\times_{[0.0,6.9]}$ & 61 & 30 \\
& Student after 2,500 steps & $1.41\!\times_{[0.0,4.6]}$ & 55 & 33 \\
\midrule
Training teachers & 55,000 & $5.20\!\times_{[1.1,12.5]}$ & 76 & 18 \\
& 1,000 & $1.67\!\times_{[0.0,5.1]}$ & 58 & 33 \\
& 500 & $1.01\!\times_{[0.0,3.8]}$ & 50 & 36 \\
& 100 & $0.00\!\times_{[0.0,2.1]}$ & 36 & 51 \\
\midrule
Backbone & ScaleGMN & $5.20\!\times_{[1.1,12.5]}$ & 76 & 18 \\
& parameter-matched GMN & $5.70\!\times_{[1.0,13.0]}$ & 76 & 18 \\
\midrule
Capacity & width 128, 4 layers & $5.20\!\times_{[1.1,12.5]}$ & 76 & 18 \\
& width 64, 4 layers & $1.83\!\times_{[0.0,5.8]}$ & 58 & 32 \\
& width 32, 4 layers & $0.69\!\times_{[0.0,3.4]}$ & 44 & 43 \\
& width 128, 3 layers & $1.58\!\times_{[0.0,5.8]}$ & 57 & 32 \\
& width 128, 2 layers & $0.62\!\times_{[0.0,3.0]}$ & 43 & 44 \\
\bottomrule
\end{tabular}}
\end{resulttable}

The communication rows isolate the role of the joint graph: direct
neuron-level cross edges outperform removing communication or replacing it by
a pooled FiLM summary, while strided and all-to-all cross-edge policies yield similar speedups.  The remaining rows show that performance depends more strongly on the
training population, message-passing depth, and width than on choosing
ScaleGMN rather than a parameter-matched GMN.  Direct teacher matching is also
more effective here than supervising the initialization toward a student
obtained after a fixed number of distillation steps.  These are single
operator runs; the table should not be read as an estimate of seed variance.

\subsection{Reach--budget plots}
\label{app:reach}

\begingroup
\renewcommand{\topfraction}{0.98}
\renewcommand{\bottomfraction}{0.98}
\renewcommand{\textfraction}{0.02}
\renewcommand{\floatpagefraction}{0.70}
\setcounter{topnumber}{4}
\setcounter{bottomnumber}{4}
\setcounter{totalnumber}{6}
\setlength{\floatsep}{10pt}
\setlength{\textfloatsep}{10pt}
\setlength{\intextsep}{10pt}
\setlength{\abovecaptionskip}{4pt}
\setlength{\belowcaptionskip}{0pt}
\makeatletter
\setlength{\@fptop}{0pt}
\setlength{\@fpsep}{10pt}
\setlength{\@fpbot}{0pt plus 1fil}
\makeatother

The speedup tables use the target $q_{\vv_T}^{\star}$ attained by each teacher's anchor. To
show progress independently of that relative target, the following figures
plot the fraction of teachers that have reached a fixed teacher-dependent
criterion by distillation budget $b$. They cover exactly the $22$ settings
reported in the main tables: all rows of
\cref{tab:inr-main,tab:classifier-main,tab:ood-task}, and the pooled rows of
\cref{tab:multi-architecture,tab:ood-architecture}.

For INRs, \emph{fidelity reach} requires
$\operatorname{PSNR}(\text{student},\text{teacher})\ge T$, while
\emph{performance-retention reach} requires
$\operatorname{PSNR}(\text{student},\text{data})
\ge\operatorname{PSNR}(\text{teacher},\text{data})-\varepsilon$.
For classifiers, fidelity is test agreement at least $T$, and retention is
student accuracy at least $(1-\varepsilon)$ times that teacher's accuracy.
For ModelNet40, we record only error to the teacher field and therefore
report fidelity reach only.

We choose thresholds to make the reach--budget plots informative: overly
easy or difficult thresholds can produce nearly flat curves near one or
zero, respectively, obscuring how reach changes with the distillation
budget.  We choose a threshold separately for each setting and criterion from
\[
T_{\rm INR}\in\{15,20,25,30,35,40\}\ \mathrm{dB},
\quad
\varepsilon_{\rm INR}\in\{.5,1,2,3,5,7,10,15,20\}\ \mathrm{dB},
\]
\[
T_{\rm clf}\in\{.95,.9,.85,.8,.75,.7,.6,.5,.4,.3\},
\quad
\varepsilon_{\rm clf}\in\{.01,.02,.05,.1,.15,.2,.3,.4,.5\}.
\]
For each setting, let $H_{\mathrm{record}}$ be the last recorded
distillation step.  We choose the strictest threshold for which
(i) the anchor reaches it for at least half the teachers by
$H_{\mathrm{record}}$, (ii) neither the anchor nor CrossGMN reaches it for
half the teachers at step $0$, and (iii) the anchor's median step to reach
it lies in $[H_{\mathrm{record}}/200,H_{\mathrm{record}}/3]$. If none qualifies, we display the most informative
available level and mark it by an asterisk.

Each curve is an empirical step function giving the fraction of teachers
whose students have reached the threshold by budget $b$.  A final value below one indicates that some students never
reach it within the recorded horizon.  Each panel shows the budget interval
in which its curves change.

We show all evaluated initializations: CrossGMN in solid blue, its anchor in
dashed orange, random initialization in dotted gray, and additional
structured baselines in thin gray.  Appendix~\ref{app:student-init}
defines the baselines, and the family-specific subsections specify which
ones are evaluated.  In thirteen of the forty panels, both CrossGMN and its
anchor eventually reach the threshold for nearly all teachers.

\begin{figure}[!htbp]
\centering
\includegraphics[width=\linewidth]{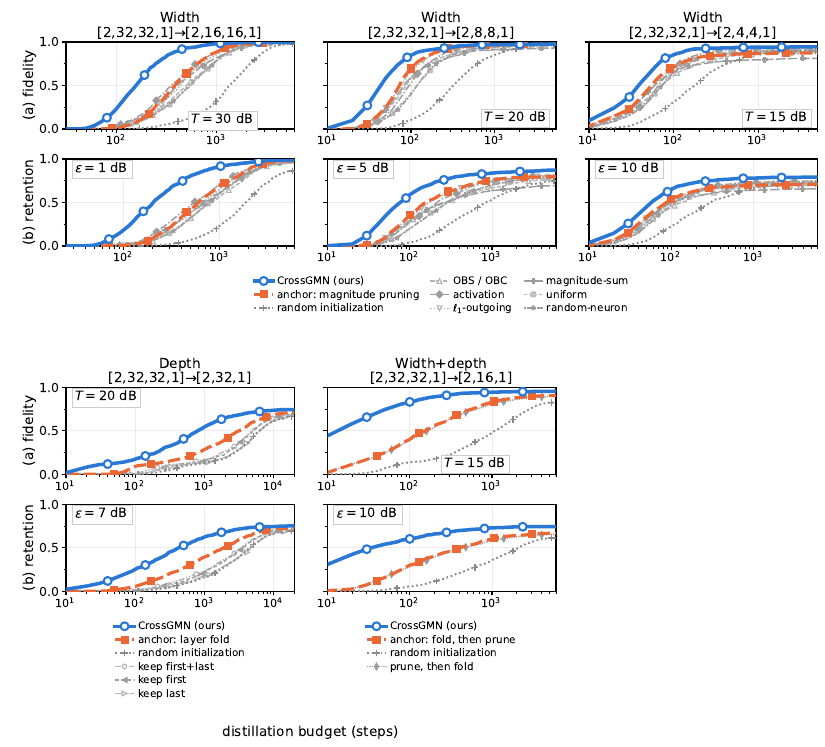}
\caption{\textbf{MNIST INR reach--budget plots} ($N=1{,}000$ per setting).
All five reductions in Table~\ref{tab:inr-main}; fidelity reach above and
performance-retention reach below.}
\label{fig:reach_app_inr_mnist}
\end{figure}

\begin{figure}[!htbp]
\centering
\includegraphics[width=\linewidth]{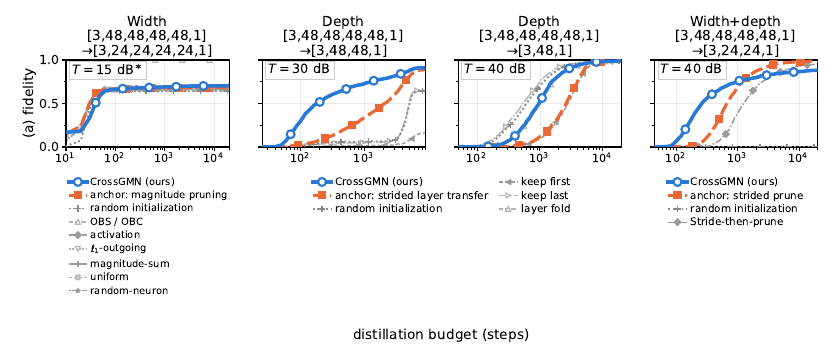}
\caption{\textbf{ModelNet40 INR reach--budget plots} ($N=1{,}000$ per setting).
All four reductions in Table~\ref{tab:inr-main}, showing fidelity to the
teacher field.  The asterisk marks a fallback threshold.}
\label{fig:reach_app_inr_modelnet}
\end{figure}

\begin{figure}[!htbp]
\centering
\includegraphics[width=\linewidth]{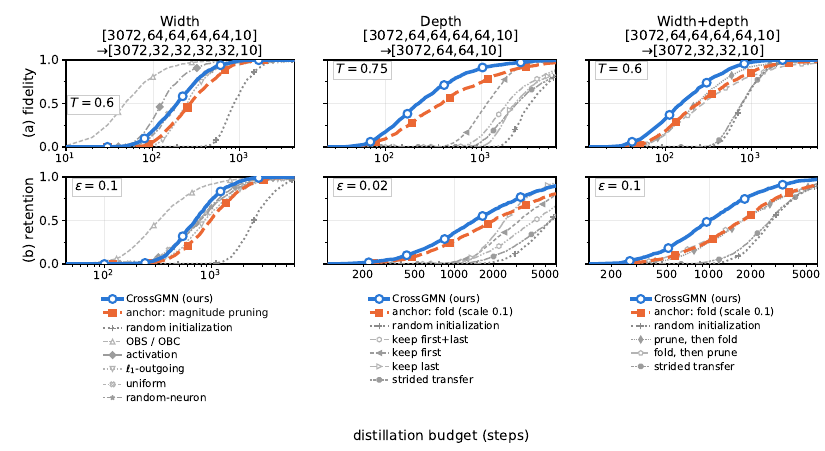}
\caption{\textbf{CIFAR-10 MLP reach--budget plots} ($N=1{,}000$ per setting).
The three reductions in Table~\ref{tab:classifier-main}; fidelity reach
above and performance-retention reach below.}
\label{fig:reach_app_clf_mlp}
\end{figure}

\begin{figure}[!htbp]
\centering
\includegraphics[width=\linewidth]{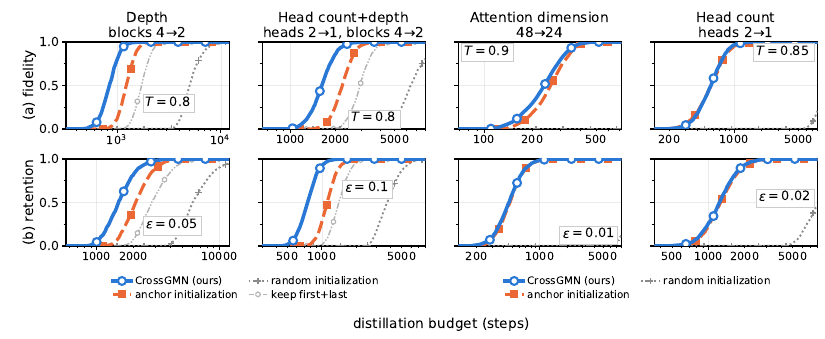}
\caption{\textbf{CIFAR-10 ViT reach--budget plots} ($N=1{,}000$ per setting).
The four reductions in Table~\ref{tab:classifier-main}; fidelity reach
above and performance-retention reach below.}
\label{fig:reach_app_clf_vit}
\end{figure}

\begin{figure}[!htbp]
\centering
\includegraphics[width=\linewidth]{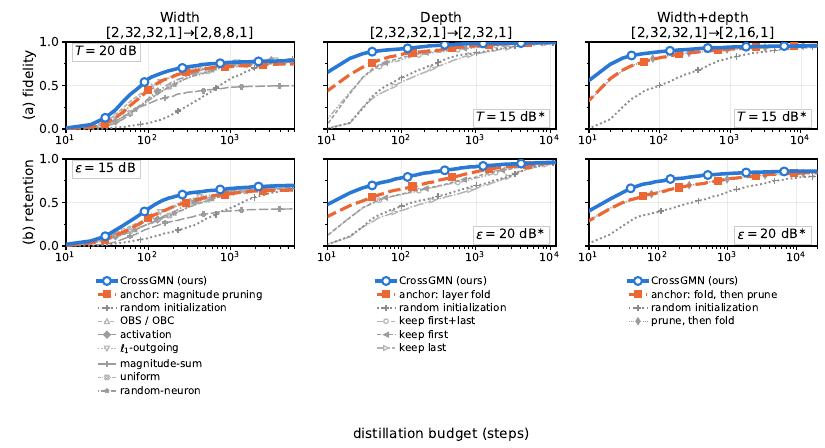}
\caption{\textbf{Transfer across datasets} ($N=1{,}000$ per setting).
MNIST-trained CrossGMN operators applied to FashionMNIST INRs without
retraining (Table~\ref{tab:ood-task}).  Asterisks mark fallback thresholds.}
\label{fig:reach_app_ood_task}
\end{figure}

\begin{figure}[!htbp]
\centering
\begin{minipage}[b]{.32\linewidth}\centering
\includegraphics[width=\linewidth]{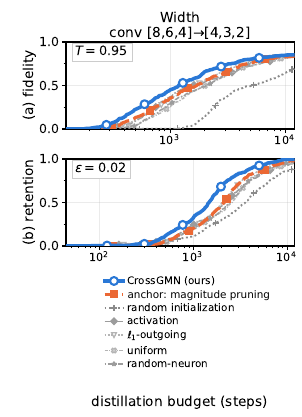}\\[2pt]
{\small (a) MNIST CNN classifiers}
\end{minipage}\hfill
\begin{minipage}[b]{.32\linewidth}\centering
\includegraphics[width=\linewidth]{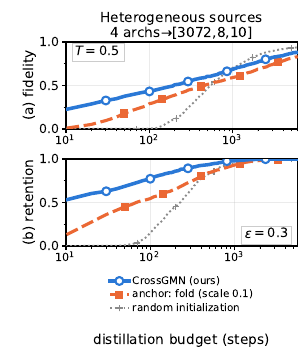}\\[2pt]
{\small (b) heterogeneous MLPs}
\end{minipage}\hfill
\begin{minipage}[b]{.32\linewidth}\centering
\includegraphics[width=\linewidth]{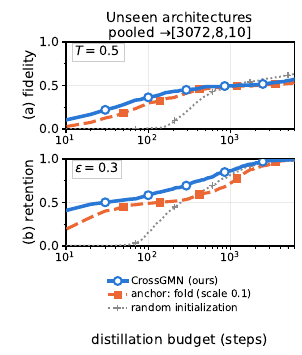}\\[2pt]
{\small (c) unseen architectures}
\end{minipage}
\caption{\textbf{Reach--budget plots for CNNs and heterogeneous MLPs.}
(a) CNN width reduction ($N=400$, Table~\ref{tab:classifier-main}).
(b) Four teacher architectures mapped to $[3072,8,10]$
($N=1{,}000$, Table~\ref{tab:multi-architecture}).
(c) The same operator on unseen source architectures
($N=500$, Table~\ref{tab:ood-architecture}).}
\label{fig:reach_app_clf_cnn}
\label{fig:reach_app_many1}
\label{fig:reach_app_ood_arch}
\end{figure}

\clearpage
\endgroup

\subsection{Complete results}
\label{app:full}

Table~\ref{tab:full} collects the complete results under the protocol in
Appendix~\ref{app:protocol}.  It reports each setting's anchor and budget,
the median and interquartile range of per-teacher speedups, the fractions
with $S_{\vv_T}>1$ and $S_{\vv_T}>2$, the failure rate, and wall-clock
speedup including initialization.  The final group is an anchor ablation.

\begin{table}[h]
\centering
\small
\caption{\textbf{Complete compression results.} Budget is $B$ from
Equation~\ref{eq:budget-B}.  Speedup statistics include failures
($S_{\vv_T}=0$).  Wall-clock speedup is the median per-teacher ratio from
Equation~\ref{eq:wall-speedup}.
$\Delta@\mathrm{Budget}$ is the median paired CrossGMN-minus-anchor quality
difference at the last evaluation step not exceeding $B$: PSNR (dB) for
INRs and test agreement for classifiers.}
\label{tab:full}
\begingroup
\setlength{\tabcolsep}{2pt}
\resizebox{\linewidth}{!}{
\begin{tabular}{@{}p{.20\linewidth}p{.18\linewidth}rrrrrrrr@{}}
\toprule
Setting & Anchor & Budget & \shortstack{Median\\speedup} & IQR & $>1$ (\%) & $>2$ (\%) & \shortstack{Failed\\(\%)} & \shortstack{Wall-clock\\speedup} & \shortstack{$\Delta@$\\Budget} \\
\midrule
\multicolumn{10}{@{}l}{\textbf{MNIST SIREN INRs}} \\
\raggedright width 3.5$\times$ & \raggedright magnitude pruning & 2,710 & $\mathbf{2.20\times}$ & $[1.24,3.86]$ & 83 & 54 & 2 & $2.17\times$ & +1.78 \\
\raggedright width 11.3$\times$ & \raggedright magnitude pruning & 2,480 & $\mathbf{5.20\times}$ & $[1.14,12.48]$ & 76 & 68 & 18 & $4.99\times$ & +1.65 \\
\raggedright width 32$\times$ & \raggedright magnitude pruning & 3,520 & $\mathbf{4.10\times}$ & $[0.00,14.10]$ & 62 & 58 & 35 & $3.98\times$ & +0.47 \\
\raggedright depth 2$\rightarrow$1 & \raggedright layer fold & 4,560 & $\mathbf{4.07\times}$ & $[2.21,6.57]$ & 90 & 78 & 7 & $4.00\times$ & +0.65 \\
\raggedright width+depth 18.2$\times$ & \raggedright fold, then prune & 2,750 & $\mathbf{8.89\times}$ & $[3.56,16.83]$ & 86 & 83 & 13 & $8.36\times$ & +0.44 \\
\midrule
\multicolumn{10}{@{}l}{\textbf{FashionMNIST SIREN INRs (transfer across datasets)}} \\
\raggedright width 3.5$\times$ & \raggedright magnitude pruning & 2,740 & $\mathbf{2.06\times}$ & $[1.15,3.16]$ & 80 & 51 & 6 & $2.04\times$ & +0.77 \\
\raggedright width 11.3$\times$ & \raggedright magnitude pruning & 2,250 & $\mathbf{2.18\times}$ & $[0.00,5.77]$ & 61 & 52 & 33 & $2.13\times$ & +0.32 \\
\raggedright depth 2$\rightarrow$1 & \raggedright layer fold & 4,320 & $\mathbf{3.09\times}$ & $[1.54,6.09]$ & 86 & 66 & 6 & $3.03\times$ & +0.32 \\
\raggedright width+depth 18.2$\times$ & \raggedright fold, then prune & 2,870 & $\mathbf{3.78\times}$ & $[1.11,10.27]$ & 76 & 64 & 17 & $3.68\times$ & +0.17 \\
\midrule
\multicolumn{10}{@{}l}{\textbf{ModelNet40 3-D INRs}} \\
\raggedright width 3.8$\times$ ($s=0.005$) & \raggedright magnitude pruning & 1,750 & $\mathbf{1.11\times}$ & $[0.88,1.73]$ & 57 & 20 & 18 & $0.98\times$ & +0.00 \\
\raggedright depth 4$\rightarrow$2 & \raggedright strided layer transfer & 4,730 & $\mathbf{6.85\times}$ & $[0.95,19.81]$ & 74 & 68 & 12 & $6.09\times$ & +3.87 \\
\raggedright depth 4$\rightarrow$1 & \raggedright strided layer transfer & 1,930 & $\mathbf{2.88\times}$ & $[1.97,4.03]$ & 96 & 74 & 0 & $2.70\times$ & +2.61 \\
\raggedright width+depth 10.1$\times$ & \raggedright strided prune & 1,830 & $\mathbf{2.46\times}$ & $[0.41,4.76]$ & 68 & 55 & 23 & -- & +0.29 \\
\midrule
\multicolumn{10}{@{}l}{\textbf{MLP (CIFAR-10)}} \\
\raggedright width 64$\rightarrow$32 & \raggedright magnitude pruning & 3,940 & $\mathbf{1.04\times}$ & $[0.75,1.35]$ & 54 & 4 & 5 & $1.01\times$ & +0.0050 \\
\raggedright depth 4$\rightarrow$2 & \raggedright layer fold (scale $0.1$) & 3,290 & $\mathbf{1.25\times}$ & $[0.73,2.27]$ & 60 & 30 & 3 & $1.15\times$ & +0.0120 \\
\raggedright width+depth & \raggedright layer fold (scale $0.1$) & 3,780 & $\mathbf{1.18\times}$ & $[0.76,1.84]$ & 59 & 21 & 3 & $1.11\times$ & +0.0100 \\
\midrule
\multicolumn{10}{@{}l}{\textbf{MNIST CNN}} \\
\raggedright conv width & \raggedright magnitude pruning & 9,900 & $\mathbf{1.17\times}$ & $[0.93,1.42]$ & 68 & 7 & 1 & $1.16\times$ & +0.0008 \\
\midrule
\multicolumn{10}{@{}l}{\textbf{ViT (CIFAR-10)}} \\
\raggedright attention dimension 48$\rightarrow$24 & \raggedright top-$k$ attention dimensions & 3,680 & $\mathbf{0.74\times}$ & $[0.63,0.85]$ & 7 & 0 & 5 & $0.73\times$ & $-$0.0011 \\
\raggedright head count 2$\rightarrow$1 & \raggedright head drop & 4,530 & $\mathbf{0.86\times}$ & $[0.78,0.96]$ & 16 & 0 & 0 & $0.85\times$ & +0.0004 \\
\raggedright depth 4$\rightarrow$2 & \raggedright strided block transfer & 6,960 & $\mathbf{1.21\times}$ & $[1.12,1.30]$ & 95 & 0 & 0 & $1.20\times$ & +0.0044 \\
\raggedright head count+depth & \raggedright strided blocks + head drop & 7,470 & $\mathbf{1.18\times}$ & $[1.09,1.27]$ & 92 & 0 & 0 & $1.16\times$ & +0.0047 \\
\midrule
\multicolumn{10}{@{}l}{\textbf{Heterogeneous-architecture compression}} \\
\raggedright pooled (four source architectures) & \raggedright layer fold (scale $0.1$) & 3,640 & $\mathbf{1.28\times}$ & $[0.71,2.25]$ & 62 & 28 & 10 & $1.20\times$ & +0.0040 \\
\midrule
\multicolumn{10}{@{}l}{\textbf{Generalization to unseen source architectures}} \\
\raggedright pooled (unseen source architectures) & \raggedright layer fold (scale $0.1$) & 3,450 & $\mathbf{1.06\times}$ & $[0.49,1.84]$ & 51 & 23 & 14 & $1.00\times$ & +0.0030 \\
\midrule
\multicolumn{10}{@{}l}{\textbf{Anchor ablation: heterogeneous-architecture compression (layer-fold reference above)}} \\
\raggedright strided layer transfer & \raggedright strided layer transfer & 4,240 & $\mathbf{1.34\times}$ & $[0.62,2.60]$ & 61 & 35 & 13 & $1.30\times$ & +0.0030 \\
\raggedright keep first+last & \raggedright keep first+last & 4,190 & $\mathbf{1.23\times}$ & $[0.54,2.36]$ & 60 & 31 & 15 & $1.20\times$ & +0.0030 \\
\raggedright random initialization & \raggedright random initialization & 3,730 & $\mathbf{0.82\times}$ & $[0.34,1.59]$ & 42 & 16 & 21 & $0.80\times$ & +0.0020 \\
\bottomrule
\end{tabular}
}
\endgroup
\end{table}

The quality difference at $B$ complements speedup: faster distillation need
not imply a better asymptotic solution.  All reported settings with median
speedup above one also have nonnegative median quality differences at this
budget.  For ViT attention-dimension and head-count reductions, the
structured anchors already provide strong initializations, and CrossGMN
does not accelerate subsequent distillation.

\subsection{Classifier teacher-population diversity}
\label{app:diversity}

As noted in Section~\ref{sec:exp_setup}, amortizing a learned compression
operator across a model population is motivated by the population spanning
diverse functions.  We therefore measure diversity in classifier teachers' outputs,
which directly characterize the functions that their students must preserve.
Parameter differences alone would be insufficient, since different weights
can realize the same function.

\paragraph{Populations and sampling.}
Table~\ref{tab:diversity} reports averages over $10{,}000$ randomly sampled
pairs of distinct teachers from each classifier population, using the full
held-out cohort evaluated for compression.  Settings that share a teacher
population share its diversity statistics.

\paragraph{Diversity metrics.}
For teachers $\vv_T$ and $\vv_T'$, let
$c_T(\vx)=\argmax_j f_{\vv_T}(\vx)_j$ and $c_{T'}(\vx)$ be their predicted
labels on a shared test set $\mathcal D$.  \emph{Prediction agreement}
measures how often their decisions coincide:
\[
p_o=\frac{1}{|\mathcal D|}\sum_{(\vx,y)\in\mathcal D}
\mathbf1[c_T(\vx)=c_{T'}(\vx)].
\]
Lower agreement indicates greater variation in predictions.  To distinguish
shared correct predictions from shared mistakes, define each teacher's
error set $E_T=\{(\vx,y)\in\mathcal D:c_T(\vx)\ne y\}$ and report
\emph{wrong-prediction intersection over union}:
\[
\operatorname{IoU}_{\mathrm{err}}(T,T')
=\frac{|E_T\cap E_{T'}|}{|E_T\cup E_{T'}|}.
\]
This measures whether the teachers fail on the same inputs, regardless of
which incorrect labels they predict.  Lower values mean less overlap in
errors, although the metric also depends on accuracy.

Finally, \emph{Cohen's $\kappa$}~\citep{cohen1960coefficient} corrects
agreement for each teacher's predicted-label frequencies:
\[
\kappa=\frac{p_o-p_e}{1-p_e},
\qquad
p_e=\sum_j p_T(j)p_{T'}(j),
\qquad
p_T(j)=\frac{1}{|\mathcal D|}\sum_{(\vx,y)\in\mathcal D}
\mathbf1[c_T(\vx)=j].
\]
Here $p_e$ is the agreement expected from independent predictions with those
label frequencies.  Thus $\kappa$ complements raw agreement by accounting
for class preferences; it does not remove all effects of accuracy or task
difficulty.  Together, the three metrics describe overall prediction
similarity, overlap in errors, and agreement beyond the label marginals.

\begin{table}[h]
\centering
\small
\caption{\textbf{Classifier teacher-population diversity.} Mean statistics
over $10{,}000$ pairs of distinct teachers.  Lower agreement, error IoU, and
$\kappa$ indicate greater diversity in their respective aspects.}
\label{tab:diversity}
\begin{tabular}{@{}lrrrr@{}}
\toprule
Teacher population & Teachers & Agreement $\downarrow$ & Error IoU $\downarrow$ & $\kappa$ $\downarrow$ \\
\midrule
CIFAR-10 MLP & 1,000 & 0.715 & 0.694 & 0.682 \\
CIFAR-10 ViT & 1,000 & 0.792 & 0.556 & 0.769 \\
MNIST CNN & 400 & 0.891 & 0.210 & 0.879 \\
CIFAR-10 MLP, four source architectures & 1,000 & 0.643 & 0.683 & 0.602 \\
\quad unseen source architectures & 500 & 0.618 & 0.653 & 0.574 \\
\bottomrule
\end{tabular}

\end{table}

\paragraph{Findings.}
On CIFAR-10, fixed-architecture MLP teachers have lower agreement than ViTs
($0.715$ versus $0.792$), also reflected in $\kappa$ ($0.682$ versus
$0.769$).  The MNIST CNNs have high agreement ($0.891$) but low error overlap
($0.210$): their predictions often coincide, while their mistakes are less
shared.  The heterogeneous MLP population and its unseen-architecture
counterpart have similar statistics, with $\kappa=0.602$ and $0.574$,
respectively.  These results establish variation within each population.

\end{document}